\documentclass[10pt,twocolumn,letterpaper]{article}
\usepackage[pagenumbers]{cvpr}

\usepackage{cuted}

\usepackage{caption}

\usepackage{microtype}
\AtBeginDocument{\setlength{\parfillskip}{0pt plus 0.62\columnwidth}}

\usepackage{graphicx}
\usepackage{booktabs}
\usepackage{adjustbox}
\usepackage{multirow}
\usepackage{amsmath,amssymb}
\usepackage{xcolor}
\usepackage{colortbl}
\usepackage{pifont}
\usepackage{tikz}
\usetikzlibrary{positioning,calc,fit,backgrounds,arrows.meta,decorations.pathreplacing,shapes.misc}
\usepackage{enumitem}
\setlist{nosep,leftmargin=1.2em}
\usepackage{xspace}

\definecolor{cvprblue}{rgb}{0.21,0.49,0.74}
\usepackage[pagebackref,breaklinks,colorlinks,allcolors=cvprblue]{hyperref}

\newcommand{\bench}{PROPS\xspace}
\newcommand{\train}{PROPS-Train\xspace}
\newcommand{\crefp}[1]{(\cref{#1})}
\newcommand{\cmark}{\textcolor{green!50!black}{\ding{51}}}
\newcommand{\xmark}{\textcolor{red!70!black}{\ding{55}}}
\newcommand{\pmark}{\textcolor{orange!80!black}{$\circ$}}
\definecolor{recall}{HTML}{4A6C94}
\definecolor{resist}{HTML}{A8693F}
\definecolor{reuse}{HTML}{557F62}
\definecolor{lookA}{HTML}{D55E00}
\definecolor{lookB}{HTML}{0072B2}
\definecolor{ours}{HTML}{FDECEC}

\title{PROPS: Can Visual Generators Keep the Current Look of a Prop?}
\hypersetup{pdftitle={PROPS: Can Visual Generators Keep the Current Look of a Prop?}}
\newcommand{\projecturl}{https://haoxiangzhao12138.github.io/PROPS/}
\newcommand{\codeurl}{https://github.com/haoxiangzhao12138/PROPS}
\newcommand{\benchurl}{https://huggingface.co/datasets/CUDAOUTOFMEMORY/PROPS-Benchmark}
\newcommand{\trainurl}{https://huggingface.co/datasets/CUDAOUTOFMEMORY/PROPS-Train}
\author{Xiangzhao Hao$^{1,2*}$\quad Dianyi Wang$^{2,3*}$\quad Keyan Zhou$^{2,3}$\quad Yiqian Zhang$^{4}$\\
Zefeng Zhang$^{5}$\quad Bailin Li$^{6}$\quad Kun Zhan$^{6}$\quad Junchi Yan$^{1,2\dagger}$\\[1.5mm]
{\normalsize $^{1}$Shanghai Jiao Tong University\quad $^{2}$Shanghai Innovation Institute\quad $^{3}$Fudan University\quad
$^{4}$Hangzhou Dianzi University}\\
{\normalsize $^{5}$Institute of Information Engineering, University of Chinese Academy of Sciences\quad
$^{6}$Li Auto Inc.}\\[1mm]
{\small\tt haoxiangzhao@sii.edu.cn, dywang24@m.fudan.edu.cn, jonaszhou01@gmail.com,}\\
{\small\tt yiqian.zyq@gmail.com, zhangzef1999@163.com, libailin@lixiang.com,}\\
{\small\tt zhankun@lixiang.com, yanjunchi@sjtu.edu.cn}\\[1.5mm]
{\small\href{\projecturl}{Project page}\quad$\cdot$\quad\href{\codeurl}{Code}\quad$\cdot$\quad
\href{\benchurl}{PROPS-Benchmark}\quad$\cdot$\quad\href{\trainurl}{PROPS-Train}}}
\hypersetup{pdfauthor={Xiangzhao Hao, Dianyi Wang, Keyan Zhou, Yiqian Zhang, Zefeng Zhang, Bailin Li, Kun Zhan, Junchi Yan}}

\begin{document}
\maketitle
{\renewcommand{\thefootnote}{}\footnotetext{$^{*}$Equal contribution.\quad $^{\dagger}$Corresponding author.}}
\begin{abstract}
Many generation tasks draw the next image from a visual history, such as storyboards, illustrated stories and
multi-turn editing. A new request often states what happens next without restating how earlier objects look.
Instead of recovering an object's current look from the history, generators often redraw it from the request alone
or bring back an outdated look, and the error can carry into later steps. We call this failure \emph{prop drift}.
Existing evaluations rarely isolate it, because similarity to the history rewards copying and a judge that sees a single history cannot tell a look recovered from that history from one the generator happens to prefer. We introduce \bench, a benchmark of 2,257 pairs of twin histories that end on the same request but show one
prop in two different looks. A pair passes only if the generator draws the matching look on each side and carries
out the request, so a look preferred by the generator or the judge passes at most one side. All 34 open and closed
systems we test show prop drift, and even systems that carry out the requested action on most sides often lose the
current look of the prop. Swapping the planners and editors of agents further shows that both history understanding and image generation limit agent performance. We also build \train, 80,250 training chains in four tracks whose targets recover an earlier state or carry scripted changes forward, and fine-tuning on it clearly improves all four tested models. Both datasets
will be publicly released.
\end{abstract}

\makeatletter\global\@topnum\z@\makeatother
\begin{figure}[t]
\centering
\includegraphics[width=\linewidth]{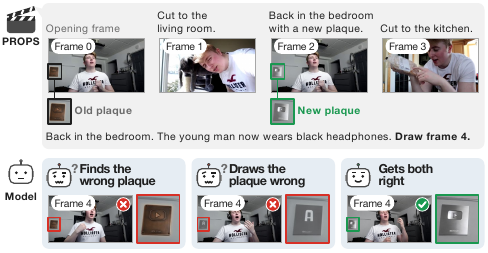}
\vspace{-5mm}
\caption{\textbf{A visual history and three model answers.} Frames alternate with shot scripts. The request says what
happens next but not how the plaque looks, which only frame 2 shows. All three add the headphones, but the
first brings back the old plaque of frame 0, the second gets the new plaque wrong, and only the third keeps it.}
\label{fig:task}
\vspace{-3mm}
\end{figure}

\newcommand\cardscaption{\textbf{\bench and \train.} Left: pairs by requirement and subtype, and the two looks of the
prop in two pairs per requirement, with Reuse crops cut from frames drawn by SenseNova-U1.5. Right: chains
by track, with intercut chains split by the number of shots cut away and synthetic chains by kind of change, and the
keyframes of one chain per track with its supervised targets marked.}
\begin{figure*}[t]
\centering
\includegraphics[width=\linewidth]{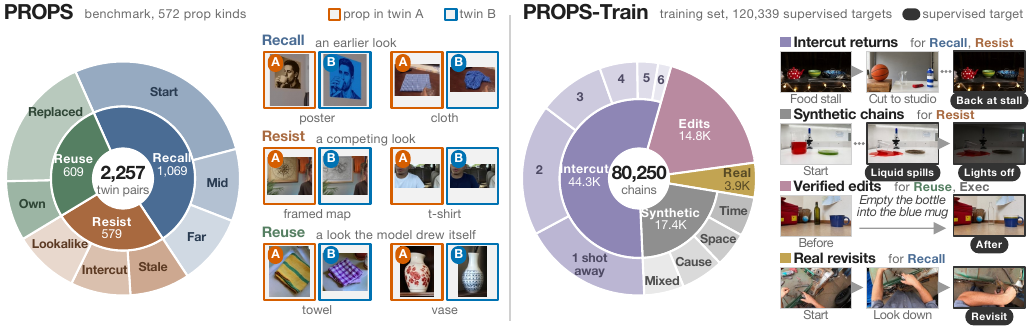}
\vspace{-5mm}
\caption{\cardscaption}
\label{fig:overview}
\vspace{-3mm}
\end{figure*}

\section{Introduction}
\label{sec:intro}

A growing range of tasks generate the next image from a visual history, including multi-turn editing, illustrated
stories, storyboards, keyframes for agentic video \cite{weave,vistorybench,dramachain,keyframecompass,vimax} and
assistants that place an object from earlier in a chat into a new scene \cite{m2ca}. In these tasks a later request
often says what happens next without restating how earlier objects look, so their current look has to come from the
earlier images and the changes made since. In \cref{fig:task} the request says only that the man now wears
headphones, whereas the plaque behind him was swapped two shots earlier and only frame 2 shows the new one. A
generator that redraws the plaque from the request alone, or brings back the old one from frame 0, fails to show the
plaque now in the story. We call it \emph{prop drift} when a generator loses the current look of a prop that the request leaves unstated. Prior work names identity preservation under state change as a key challenge when generating from such a history \cite{icgsurvey}. Prop drift is one failure of this kind, and it can carry forward, since each generated image
enters the history of later steps.

Existing evaluations rarely isolate prop drift and mostly report per-image fidelity, prompt alignment or generic
subject consistency \cite{geneval,vbench,vistorybench}. Recent benchmarks add visual memory, revisits and hidden state
\cite{vmembench,r2mbench,statealias,worldstate}, yet they seldom fix which earlier image defines the current look, and few check that copying a history frame and reading only the script both fail \crefp{tab:compare}. Without that constraint, similarity to the history rewards copying, and on 152 pairs CLIP, SigLIP, DINOv2 and DreamSim \cite{clip,siglip,dinov2,dreamsim} all rank a copy of the opening frame above the ground truth and the outputs of four generators. A VLM judge that sees only one history cannot tell a look taken from that history from one that the generator happens to prefer, and the judge may prefer the same look. Given the same outputs paired with either their own history or a history showing the prop in another look,
three VLM judges, Gemma-4, Qwen3.8 and GPT-5.5, score the matching history higher on only 52 to 61\% of sides, against 50\% by chance \crefp{sec:meta}.

We introduce \bench, a benchmark that measures prop drift with \emph{twin histories} \crefp{fig:twins}. Each side
gives a generator the keyframes and shot scripts of a storyboard up to a query step, then the query script, and the
generator draws the next keyframe. The two twins end on the same query step and differ only in the current look of
one prop. Each look is thus right on one side and wrong on the other, so a look that the generator or the judge
prefers passes at most one side. When no script describes the look, the twins also share
every script, so a generator that follows only the scripts or copies a shared frame gives the same output on both
sides and cannot match both looks.  For
\textbf{Follow} a judge decides which of the two looks each output shows, if either, and for \textbf{Exec} it
answers yes or no questions about the scripted action. Neither check needs a target image or an open-ended rating \cite{mllmjudge,viescore}, and a pair passes only if both sides pass both. The 2,257 pairs come from 11 video sources and synthetic
chains and cover eight subtypes of three requirements \crefp{fig:overview,fig:subtypes-def}.
\textbf{\textcolor{recall}{Recall}} recovers a look shown several images earlier.
\textbf{\textcolor{resist}{Resist}} selects the current look when the history also holds a stale version, an
intercut storyline or a look-alike object. \textbf{\textcolor{reuse}{Reuse}} keeps a look the generator drew itself
from a script that names the look once.

Most targets in keyframe chains cut from videos can be drawn from the previous frame, so training on such chains
mainly teaches a model to continue the last image. We therefore build \train, 80,250 chains with 120,339 targets that recover an earlier state or carry scripted changes forward, in four tracks for the same three requirements
\crefp{fig:overview}. Real revisits return to an earlier place for Recall, and intercut returns go back to a storyline after shots of another one for Recall and Resist. Synthetic chains carry two to five changes in time, space or cause, so that an outdated look is always wrong, and verified edits make one scripted change and keep everything else for Reuse and Exec. 

We evaluate 34 systems in four families: unified multimodal models, image generators and editors,
agents in which a language model plans and an editor draws, and keyframe-conditioned video models. All of them show prop drift, and open systems show it most. On all 2,257 pairs the best open system before fine-tuning, FLUX.3-klein, scores
30.4\%, and 20 of the 25 open systems stay below 10\%. On a shared 600-pair split closed image generators score 39.3 to 50.4\% and agents with closed planners reach 71.0\%, so the best system still falls 29.0 points short
\crefp{tab:closed,fig:follow-exec}. On this split SenseNova-U1.5 carries out the scripted action on 88.5\% of sides but keeps the current look on only 47.6\%. Agents let us vary understanding and generation one at a time. Starting
from Qwen3.8 planning for Qwen-Image-Edit at 19.1\%, the stronger GPT-5.6 planner raises the agent to 55.7\%, the
stronger GPT Image 2.5 editor raises it to 29.3\%, and only with both does it reach 71.0\%. Both abilities thus limit
the score, and for this open agent understanding is the larger gap.  On the full benchmark, \train lifts SenseNova-U1.5 from 26.7\% to 47.0\%, mostly through Follow.

\noindent Our contributions are:
\begin{itemize}
  \item \textbf{\bench}, a benchmark of 2,257 twin pairs across eight subtypes of Recall, Resist and Reuse that
    measures prop drift without a target image, built by editing one prop into two looks in the same real or synthetic keyframes.
  \item \textbf{\train}, 80,250 chains in four tracks whose targets recover earlier states or carry scripted changes forward. Fine-tuning on it clearly improves all four tested models.
  \item \textbf{An evaluation of 34 open and closed systems} across four families. All of them show prop drift, and
    carrying out the scripted action does not mean keeping the look. In agents, swapping planners and editors shows that both history understanding and image generation limit the score.
\end{itemize}

\begin{table}[t]
\centering
\scriptsize
\setlength{\tabcolsep}{3.2pt}
\captionsetup{position=top}
\caption{\textbf{Comparison with related benchmarks.} Multi-shot items are shot sequences of one story, and Image history gives several images as input. Own output scores later steps on content the model drew itself. Shortcuts fail means that the paper reports near-zero scores both for copying a history frame and for reading only the script, which for \bench holds on Recall and Resist but not on Reuse. Train set means that a training set is or will be released. \pmark\ marks partial support, and sources are  in \cref{app:tab1}.}
\label{tab:compare}
\begin{tabular}{@{}lccccc@{}}
\toprule
 & Multi- & Image & Own & Short- & Train \\
Benchmark & shot & history & output & cuts fail & set \\
\midrule
ViStoryBench \cite{vistorybench}         & \cmark & \pmark & \pmark & \xmark & \xmark \\
DramaChain \cite{dramachain}             & \cmark & \pmark & \pmark & \xmark & \xmark \\
KeyFrame-Compass \cite{keyframecompass}  & \cmark & \cmark & \xmark & \xmark & \xmark \\
LongMIG-Bench \cite{longmig}             & \cmark & \pmark & \cmark & \xmark & \pmark \\
VMem-Bench \cite{vmembench}              & \cmark & \pmark & \cmark & \xmark & \xmark \\
WEAVE \cite{weave}                        & \pmark & \cmark & \pmark & \xmark & \cmark \\
MSE-Bench (VINCIE) \cite{vincie}         & \pmark & \cmark & \cmark & \xmark & \cmark \\
OCCUR-Bench \cite{respec}                 & \pmark & \cmark & \pmark & \xmark & \xmark \\
IntReal \cite{intreal}                    & \cmark & \cmark & \xmark & \xmark & \cmark \\
StEvo-Bench \cite{stevo}                  & \xmark & \pmark & \pmark & \xmark & \xmark \\
WROP \cite{wrop}                         & \xmark & \cmark & \xmark & \xmark & \cmark \\
MemoBench \cite{memobench}                & \xmark & \pmark & \pmark & \xmark & \xmark \\
R2M-Bench \cite{r2mbench}                 & \xmark & \pmark & \cmark & \xmark & \xmark \\
VBVR-Pro \cite{vbvrpro}                   & \pmark & \pmark & \pmark & \pmark & \cmark \\
StateAliasBench \cite{statealias}         & \xmark & \cmark & \xmark & \xmark & \cmark \\
WorldState-Bench \cite{worldstate}        & \xmark & \pmark & \xmark & \xmark & \xmark \\
\midrule
\rowcolor{ours}\bench (ours)             & \cmark & \cmark & \cmark & \pmark & \cmark \\
\bottomrule
\end{tabular}

\vspace{-3mm}
\end{table}

\section{Related Work}
\label{sec:related}

\paragraph{Generation from a visual history.}
Many generators condition the next image on earlier ones. Story visualization keeps characters consistent across
one image per sentence \cite{storygan,storydiffusion}, and UniMAGE \cite{unimage} writes a script and its keyframes in
one model. Unified multimodal models \cite{bagel,emu35,omnigen2,januspro,showo2,sensenova} and in-context editors
\cite{vincie,fluxkontext,qwenimage} take interleaved histories for multi-turn editing, and video models turn
keyframes into clips \cite{wan,ltxvideo}. For long histories, StoryMem \cite{storymem} conditions each new shot on a memory bank of earlier keyframes and ViMax \cite{vimax} plans long videos with agents that track characters and settings across scenes. Other agents select reference images from a conversation and call an editor \cite{dramachain,m2ca,qwenimageagent}, and M2CA \cite{m2ca} finds that retrieving the right visual context is their main bottleneck. Visual instructions must show the state that earlier steps left \cite{showhowto,liger,instructioncrafter},
reasoning with generated images builds on states drawn before \cite{mvot,deltav,see2think}, and neural game engines
and world models predict new observations from earlier ones \cite{gamengen,memworld}. In each setting a returning object should keep the look its history implies, including any change made since it was last shown, and a wrong look carries into later steps.

\paragraph{Benchmarks for consistency over a history.}
Story benchmarks evaluate character identity, style and prompt alignment across shots \cite{vistorybench,dramachain},
while KeyFrame-Compass \cite{keyframecompass} tests whether video generators reproduce given keyframes. Interleaved
generation benchmarks either score each image on its own \cite{geneval,realunify,unimmmu,envision} or rate the full
output with an MLLM judge \cite{opening,isgbench,interleavedbench,comm}. IntReal \cite{intreal} tests next-image
generation under real histories, and multi-reference and long-context suites show that scores fall as references or
history grow \cite{micon,multibanana,unilonggen}. Multi-turn editing benchmarks likewise find that unedited content
drifts over turns \cite{weave,omniiie,compbench,mtedit}, and OCCUR-Bench \cite{respec} asks an editor to restore
content that a later edit hid. Closer to our setting, world-model and long-horizon benchmarks check revisited places,
hidden state that evolves out of view and changed objects that reappear
\cite{worldscore,stevo,statebench,wrop,worldmembench,memobench,r2mbench}. WorldCast \cite{worldcast}, ImageTime
\cite{imagetime} and VBVR-Pro \cite{vbvrpro} check successive states against stage-wise rules, while LongMIG-Bench
\cite{longmig}, VMem-Bench \cite{vmembench}, EvoState \cite{evostate} and StoryEngine \cite{storyengine} add revisits,
stale looks and look-alikes to long stories.

\paragraph{Shortcut analysis and controlled pairs.} Several studies find that standard consistency metrics are
vulnerable to shortcuts that ignore the history. In ShowHowTo \cite{showhowto} copying the input image beats the
evaluated model on scene consistency, repeating one image maximizes the consistency score of InstructionCrafter
\cite{instructioncrafter}, and most multi-turn tests can be solved from the last image \cite{nonmarkov}. Other work
rules such shortcuts out by design. Minimal pairs \cite{blimp,contrastsets,winoground} and counterfactual
interventions \cite{cronos} vary one factor so that a correct answer must depend on it, video understanding
benchmarks drop items that can be answered without temporal evidence \cite{tocbench}, and the concurrent
StateAliasBench \cite{statealias} and WorldState-Bench \cite{worldstate} pair histories that end on the same action
but differ in hidden state. \bench brings paired-history control to the look of a prop in a visual history and adds
an execution check, so that neither the script alone nor a copy of a history frame can pass a Recall or Resist pair.

\section{The \bench Benchmark}
\label{sec:bench}
\begin{figure*}[t]
\centering
\includegraphics[width=\linewidth]{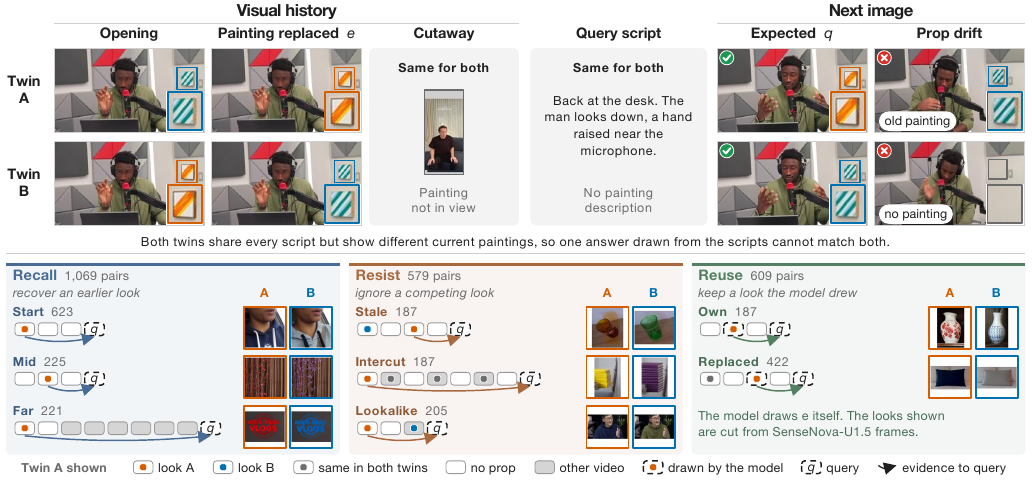}
\vspace{-5mm}
\caption{\textbf{Prop drift and twin histories.} Top: a Resist-Stale pair. The twins share every script and differ
only in how the painting looks, so one answer drawn from the scripts cannot match both. SenseNova-U1.5 keeps the old painting in twin A and draws no painting in twin B. Bottom: the history of twin A in each subtype, with the evidence frame
linked to the query, and the two looks of one pair. The Reuse looks shown come from SenseNova-U1.5, and each system is scored on its own $e$. Numbers count pairs.}
\label{fig:method}\label{fig:twins}\label{fig:subtypes-def}
\vspace{-3mm}
\end{figure*}

\bench tests whether a generator keeps the current look of a prop that the query script leaves unstated. Such a test needs three properties. A look that the generator or the judge prefers must not pass, an output that ignores the evidence must fail, and scoring must not need a target image, since a correct frame need not match the recorded one. Twin histories provide all three \crefp{sec:task}, and the construction enforces them on every Recall and Resist pair \crefp{sec:construction}.

\subsection{Task design}
\label{sec:task}\label{sec:twins}\label{sec:control}\label{sec:subtypes}
A storyboard chain is a sequence of keyframes $I_0,\dots,I_L$ with shot scripts $s_1,\dots,s_L$, where $s_t$ says
only what happens between $I_{t-1}$ and $I_t$. At a query step $q$ the generator receives the interleaved history
$H_q=(I_0,s_1,I_1,\dots,s_{q-1},I_{q-1},s_q)$ and draws $\hat I_q$. Some history frames may have been drawn by the model itself. A designated prop $X$ has a \emph{current look}, the look it has in the \emph{evidence frame} $e<q$, the last frame before $q$ that shows $X$ at its place, and $s_q$ does not restate it. \emph{Prop drift} occurs when $X$ in $\hat I_q$ does not show this look.

\paragraph{Twin histories.} Each pair has two histories, twin A and twin B, that end on the same query step and give
$X$ two different looks \crefp{fig:twins}. A generator that uses its history must draw look A after history A and look B after
history B. In Recall and Resist the twins share every script, so an output drawn from the scripts alone or copied
from a frame that both twins share is the same on both sides and cannot match both looks. In Reuse the twins are given the same frames and differ only in the script of $e$, which names a different look on each side. The generator draws $e$ itself, so the look to keep is the one it drew. In every subtype the right answer is one of two known looks, so scoring needs no target image, and since each look is right on one side and wrong on the other, a look that the generator or the judge prefers passes at most one side \crefp{app:look}.

\paragraph{Three requirements.} A generator can miss the evidence because it lies far back, because the history
also shows a competing look, or because the generator drew the evidence itself. The three requirements test these
cases, and each of their eight subtypes fixes one history layout \crefp{fig:method}.
\textbf{\textcolor{recall}{Recall}} tests recovery from a frame several shots back. The evidence is the opening frame in \emph{Start}, a middle keyframe in \emph{Mid}, and again the opening frame in \emph{Far}, where frames cut in from other chains stretch the gap to eight shots.
\textbf{\textcolor{resist}{Resist}} tests whether the valid look wins over a competing one. In \emph{Stale} the opening frame shows the other twin's look and a later script says that $X$ was replaced. In \emph{Intercut} the chain alternates with another video, the same in both twins and often showing an object of the same kind, and in \emph{Lookalike} an inserted frame shows an object of the same kind in the other twin's look. \textbf{\textcolor{reuse}{Reuse}} tests whether the generator
keeps a look that it drew itself at $e$. In \emph{Own} the script of $e$ names the look once. In \emph{Replaced} both
twins open on the same unedited prop, and two steps before $q$ the generator draws the replacement from a script that
names the new look. The generator may keep this look through its own image or through the text, as a shot script names a newly introduced prop. A text-only pipeline that writes one prompt from the scripts thus passes 42.6\% of Reuse pairs on a fixed set of 600 pairs, but 0.4\% of Recall and 0.0\% of Resist pairs \crefp{app:build}.

\subsection{Scoring}
\begin{figure*}[t]
\centering
\includegraphics[width=\linewidth]{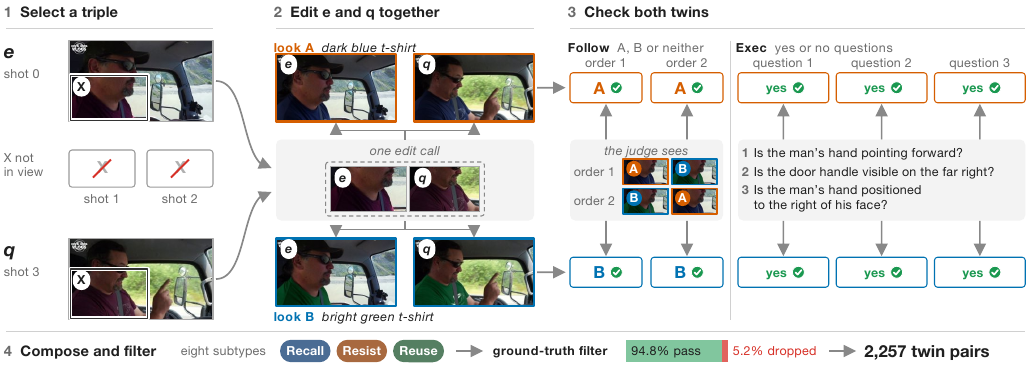}
\vspace{-1mm}
\caption{\textbf{Building \bench, shown on one Recall-Start pair.} \textbf{1} The t-shirt $X$ is seen in
$e$ and $q$ but not in between. \textbf{2} One edit call changes $X$ in both frames, once per look. \textbf{3} On the true $q$ of both twins the judge must pick the right look in both orders and answer every kept question about $s_q$.
\textbf{4} Each pair is laid out as one subtype, and the 5.2\% of candidates whose true frames fail are dropped.}
\label{fig:pipeline}
\vspace{-2mm}
\end{figure*}

\label{sec:scoring}
Each side is scored on two checks. For \textbf{Follow} a judge \cite{gemma4} sees the two reference crops and $\hat
I_q$ and answers A, B or neither. We ask twice with the crops swapped, and the side passes only if both answers name
its own look. In Reuse the reference crops are cut from the frame $e$ that the generator drew, where Qwen3.8 \cite{qwen38} grounds $X$, and a side fails if $X$ cannot be found there. For \textbf{Exec} a VLM writes yes or no \emph{execution questions} about the change in $s_q$, for example whether the fridge door is open. A query keeps 2.2 of them on average after the checks of \cref{sec:construction}, and every one must be answered correctly on $\hat I_q$.

A pair passes only if both sides pass both checks, and a missing or refused output fails its side. A copy of a frame
that differs between the twins can show the right look, but it does not carry out $s_q$ and fails Exec, so copying frame 0 scores 0.1\% and copying the previous frame scores 0.0\%.
A subtype's score is the percentage of its pairs that pass, the score of each requirement is the mean of its subtypes, and Overall is the mean of the three requirements, so that the 623 Recall-Start pairs do not dominate. Follow and Exec rates over sides serve only as diagnostics, and \cref{sec:judge} checks the judge against human annotation.

\subsection{Construction}
\label{sec:construction}\label{sec:stats}
Pairs are built from keyframe chains, in which a scene-change detector picks the keyframes and each shot script states only what changed. Egocentric video from Ego4D \cite{ego4d}, Ego-Exo4D \cite{egoexo4d}, EPIC-KITCHENS \cite{epic}, EgoLife \cite{egolife} and three smaller egocentric sets contributes 1,066 pairs, the long-video datasets Vript \cite{vript}, Video-MME \cite{videomme} and MLVU \cite{mlvu} contribute 617, and YouTube vlogs contribute 306. The other 268 pairs come from synthetic chains that return to a place by editing its real keyframe to another time of day or weather. Each candidate then passes the four steps of \cref{fig:pipeline}.

\paragraph{1. Select a triple.} A triple $(X,e,q)$ qualifies when $X$ is clearly visible in $e$, $q$ returns to the
place of $X$, no frame of that place in between shows $X$, and no script up to $q$ describes its look, so the look of
$X$ can come only from $e$. We then ground $X$ again in $e$ and $q$ to cut its reference crops.

\paragraph{2. Edit $e$ and $q$ together.} Qwen-Image-Edit-2511 \cite{qwenimage} sees crops around $X$ from $e$ and $q$ side by side and changes $X$ in both crops into look A with one call, then into look B with another. Each look keeps the kind of $X$ and its place in the frame and changes its appearance, such as its colour, pattern or printed content. Everything outside an extended SAM\,2.1 \cite{sam2} mask around $X$ is pasted back from the source frame, so the pixels of the twins differ only inside this mask \crefp{app:build}.

\paragraph{3. Check both twins.} Given the crops of A and B in both orders, the judge must pick the right look on the
true $q$ of each twin. We keep an execution question only if it is the same for both twins, does not mention $X$, is
answered correctly on both true frames, and cannot be answered from any earlier same-place frame or without $s_q$,
so copying a frame or ignoring the script fails Exec.

\paragraph{4. Compose and filter.} Each checked triple is laid out as the history of one of the eight subtypes
\crefp{app:build}. A last filter scores the true frames of every candidate with the protocol of \cref{sec:scoring} and
drops the 5.2\% that fail.

\paragraph{Statistics.} \bench has 2,257 pairs and 4,514 sides, at least 187 per subtype, over 572 prop kinds. The gap between evidence and query ranges from 2 to 8 shots \crefp{fig:stats}.

\section{\train}
\label{sec:train}
\train keeps only steps that bring back an earlier state or carry scripted changes forward. Unlike \bench, it has no twins, and each target is a full frame rather than
one edited object. It has 80,250 chains with 293K steps over 132K distinct images, since intercut returns are spliced from the frames of synthetic chains. These chains give 120,339 supervised targets in four tracks, 56.1K in synthetic chains, 42.1K in intercut returns, 14.0K in verified edits and 8.1K in real revisits \crefp{fig:overview,fig:train}. 

\paragraph{Tracks.} Generators fail the three requirements of \cref{sec:subtypes} for different reasons, so each track supplies the targets that one or two of them need.
\textbf{Real revisits} train \textcolor{recall}{Recall} on video groups that \bench does not use. A target is
supervised only when the camera returns to an earlier place and the frame looks more like that earlier view than
like the previous frame under DINOv2 \cite{dinov2}. \textbf{Intercut returns} train \textcolor{recall}{Recall} and
\textcolor{resist}{Resist}. Each splices two synthetic chains, cutting away from chain A for one to six shots of chain B, and returns with a script that names the place of A again, for example \emph{Returning to the breakfast table}, so the model has to recover the state of A instead of continuing chain B. \textbf{Synthetic chains} train \textcolor{resist}{Resist}. Their scripts make two to five changes in time, space, cause or a combination, and every later target must keep all earlier changes, so an outdated state is always
wrong. \textbf{Verified edits} train \textcolor{reuse}{Reuse} and Exec. Each step applies one scripted change and keeps everything else, including earlier additions.

\paragraph{Construction.} A synthetic script combines a sampled place and theme with one sampled kind of change
per step and never names the look of an object already in view. Qwen-Image-2.1 \cite{qwenimage} draws frame 0 and edits each later step, and Qwen3.8 \cite{qwen38} verifies every kept step, which must show the scripted change and keep the rest. \bench is scored by a different judge, Gemma-4
\cite{gemma4}, and no judge score enters the training loss. Training data and benchmark share no source video group, and removing the only two benchmark pairs with an image of DINOv2 similarity above 0.95 to a training image leaves every Overall in \cref{tab:sft} unchanged at one decimal \crefp{app:disjoint}.

\renewcommand{\dblfloatpagefraction}{0.85}
\section{Experiments}
\label{sec:exp}

\subsection{Setup}
\label{sec:setup}
\paragraph{Systems.} We run 25 open systems on the full benchmark \crefp{tab:main}, listed in \cref{tab:mainfull}:
nine unified multimodal models, eight image generators and editors, two agents with an open planner, and six keyframe-conditioned video models, scored on the last frame of each
shot. Nine closed systems run on phase~A, a fixed random draw of 600 pairs with 75 per subtype, and are compared with
open systems only there \crefp{tab:closed}. They are three image generators and six agents with a closed planner, four with Qwen-Image-Edit and two with another editor, for 34 systems in all. On phase~A we also run six agent variants for analysis \crefp{app:agents}. Every system receives the full history $H_q$ in its native input format. A system whose fixed image cap is shorter than the history receives frame 0 and the latest frames, or only the previous frame if it takes one image. The three-image models FireRed-Edit, Qwen-Image-Edit and OmniGen2 stay below 2\% even on the 2,118 pairs whose evidence frame they receive \crefp{app:models}. Each system draws its frames once with a fixed seed, and Gemma-4-31B \cite{gemma4} scores them. Intervals are 95\% cluster bootstraps over source videos, about $\pm$3 points for Overall on the full benchmark and $\pm$4.5 on phase~A.

\paragraph{Fine-tuning.} We fully fine-tune SenseNova-U1.5 \cite{sensenova} on \train at a learning rate of $2\times10^{-6}$ and BAGEL \cite{bagel} at $10^{-5}$, both for 2,000 steps on the full interleaved history and evaluated with EMA weights. Qwen-Image-2.1 and Qwen-Image-Edit are fine-tuned with LoRA of rank 64 for 4,000 and 2,000 steps, the latter on a 32K subset of the targets with the same share of each track \crefp{app:ft}.

\begin{table*}[tp]
\centering
\scriptsize
\setlength{\tabcolsep}{3.2pt}
\captionsetup{position=top}
\caption{\textbf{Open systems on the full \bench}, pass rates of pairs per subtype and their means in \%, with Follow and Exec over sides and the best value of each family in bold. The other open systems, all below 2\% \crefp{tab:mainfull}, are HiDream-O1 \cite{hidreamo1}, LLaDA2-Uni \cite{llada2uni}, OmniGen2 \cite{omnigen2}, Lumina-DiMOO \cite{lumina}, Janus-Pro \cite{januspro}, FireRed-Edit \cite{firered}, Qwen-Image-Edit \cite{qwenimage}, Wan-VACE \cite{vace}, StoryMem \cite{storymem}, HunyuanVideo-I2V \cite{hunyuanvideo} and Wan-TI2V \cite{wan}.}
\label{tab:main}
\begin{tabular}{@{}l c ccc ccc ccc cc cc@{}}
\toprule
 & & \multicolumn{3}{c}{Groups} & \multicolumn{3}{c}{\textcolor{recall}{Recall}} & \multicolumn{3}{c}{\textcolor{resist}{Resist}} & \multicolumn{2}{c}{\textcolor{reuse}{Reuse}} & \multicolumn{2}{c}{Diagnostic} \\
\cmidrule(lr){3-5}\cmidrule(lr){6-8}\cmidrule(lr){9-11}\cmidrule(lr){12-13}\cmidrule(l){14-15}
System & Overall & Recall & Resist & Reuse & Start & Mid & Far & Stale & Intercut & Lookalike & Own & Replaced & Follow & Exec \\
\midrule
\multicolumn{15}{@{}l}{\textit{Unified multimodal models}} \\
SenseNova-U1.5 \cite{sensenova} & \textbf{26.7} & \textbf{39.4} & \textbf{27.9} & \textbf{12.9} & \textbf{45.9} & \textbf{31.1} & \textbf{41.2} & 2.1 & \textbf{44.9} & \textbf{36.6} & \textbf{23.5} & 2.4 & 46.3 & \textbf{87.8} \\
Emu3.5 \cite{emu35} & 15.9 & 19.6 & 15.7 & 12.3 & 24.9 & 19.1 & 14.9 & \textbf{9.6} & 20.9 & 16.6 & 12.3 & \textbf{12.3} & \textbf{55.4} & 57.0 \\
BAGEL \cite{bagel} & 4.0 & 4.9 & 2.9 & 4.3 & 7.2 & 5.3 & 2.3 & 2.1 & 2.7 & 3.9 & 2.7 & 5.9 & 21.5 & 60.5 \\
Show-o2 \cite{showo2} & 0.0 & 0.0 & 0.0 & 0.0 & 0.0 & 0.0 & 0.0 & 0.0 & 0.0 & 0.0 & 0.0 & 0.0 & 1.4 & 11.9 \\
\midrule
\multicolumn{15}{@{}l}{\textit{Image generators and editors}} \\
FLUX.3-klein \cite{flux3klein} & \textbf{30.4} & \textbf{38.1} & \textbf{29.4} & \textbf{23.6} & \textbf{42.9} & \textbf{33.8} & \textbf{37.6} & \textbf{8.6} & \textbf{41.7} & \textbf{38.0} & \textbf{29.4} & \textbf{17.8} & \textbf{55.1} & \textbf{84.2} \\
FLUX.2-klein \cite{flux2} & 3.1 & 2.5 & 1.7 & 5.1 & 4.5 & 3.1 & 0.0 & 1.1 & 1.1 & 2.9 & 5.9 & 4.3 & 16.3 & 49.4 \\
VINCIE \cite{vincie} & 2.9 & 2.9 & 4.0 & 1.8 & 6.9 & 1.3 & 0.5 & 3.7 & 4.8 & 3.4 & 0.5 & 3.1 & 18.9 & 40.8 \\
Qwen-Image-2.1 \cite{qwenimage} & 2.2 & 2.0 & 1.9 & 2.8 & 3.2 & 2.7 & 0.0 & 0.5 & 3.7 & 1.5 & 4.3 & 1.4 & 11.4 & 51.8 \\
FLUX.2-dev \cite{flux2} & 2.0 & 1.6 & 1.4 & 2.9 & 2.6 & 2.2 & 0.0 & 1.1 & 1.6 & 1.5 & 4.8 & 0.9 & 10.5 & 49.9 \\
FLUX.1-Kontext \cite{fluxkontext} & 0.0 & 0.0 & 0.0 & 0.0 & 0.0 & 0.0 & 0.0 & 0.0 & 0.0 & 0.0 & 0.0 & 0.0 & 2.5 & 21.9 \\
\midrule
\multicolumn{15}{@{}l}{\textit{Agents (open planner + editor)}} \\
Qwen3.8 + Qwen-Edit & \textbf{19.6} & \textbf{20.6} & \textbf{14.1} & \textbf{24.2} & \textbf{22.5} & 27.6 & \textbf{11.8} & \textbf{18.2} & \textbf{9.6} & \textbf{14.6} & 31.0 & \textbf{17.5} & \textbf{38.4} & \textbf{83.1} \\
Qwen3.8 + Qwen-Image-2.1 & 16.9 & 17.0 & 10.8 & 23.0 & 13.2 & \textbf{30.7} & 7.2 & 17.6 & 5.9 & 8.8 & \textbf{32.6} & 13.3 & 35.2 & 79.6 \\
\midrule
\multicolumn{15}{@{}l}{\textit{Keyframe-conditioned video models}} \\
LTX-2.5 \cite{ltxvideo} & \textbf{8.3} & \textbf{12.1} & \textbf{8.9} & \textbf{4.0} & \textbf{15.4} & \textbf{13.3} & \textbf{7.7} & \textbf{5.9} & \textbf{10.2} & \textbf{10.7} & \textbf{4.3} & \textbf{3.8} & \textbf{30.9} & \textbf{60.8} \\
Wan-I2V \cite{wan} & 0.0 & 0.0 & 0.0 & 0.1 & 0.0 & 0.0 & 0.0 & 0.0 & 0.0 & 0.0 & 0.0 & 0.2 & 3.2 & 43.9 \\
\midrule
\multicolumn{15}{@{}l}{\textit{Fine-tuned on \train (ours)}} \\
\rowcolor{ours}SenseNova-U1.5 + \train & \textbf{47.0} & \textbf{48.3} & \textbf{52.0} & \textbf{40.6} & \textbf{55.1} & \textbf{33.8} & \textbf{56.1} & \textbf{51.3} & \textbf{48.1} & \textbf{56.6} & \textbf{32.1} & \textbf{49.1} & \textbf{71.2} & \textbf{86.1} \\
\rowcolor{ours}BAGEL + \train & 14.9 & 16.9 & 15.1 & 12.6 & 25.0 & 11.1 & 14.5 & 11.2 & 16.6 & 17.6 & 9.1 & 16.1 & 41.9 & 69.7 \\
\rowcolor{ours}Qwen-Image-2.1 + \train (LoRA) & 11.2 & 11.2 & 13.6 & 8.9 & 20.5 & 6.7 & 6.3 & 15.0 & 10.2 & 15.6 & 4.3 & 13.5 & 50.1 & 45.7 \\
\rowcolor{ours}Qwen-Image-Edit + \train (LoRA) & 10.8 & 12.8 & 12.3 & 7.2 & 25.0 & 7.6 & 5.9 & 8.6 & 16.0 & 12.2 & 3.7 & 10.7 & 52.4 & 38.7 \\
\bottomrule
\end{tabular}
\vspace{-3mm}
\end{table*}

\subsection{Benchmarking}
\label{sec:main}\label{sec:diag}\label{sec:counter}
\paragraph{Prop drift is widespread.} Among open systems without fine-tuning, FLUX.3-klein scores 30.4\% and
SenseNova-U1.5 26.7\%, and only three others pass more than 10\% \crefp{tab:main}. Strong text-to-image models also
score low, Qwen-Image-2.1 at 2.2\% and FLUX.2-dev at 2.0\%, and they keep the current look on only about 11\% of sides. Closed systems do better but do not solve the benchmark, with up to 71.0\% for agents with closed planners and 50.4\% for Nano Banana~2 on phase~A \crefp{tab:closed}.

\begin{table}[t]
\centering
\scriptsize
\setlength{\tabcolsep}{1.9pt}
\captionsetup{position=top}
\caption{\textbf{Closed systems and agents} on the 600 pairs of phase~A. $e$ shown is the \% of sides with a stored plan on which the planner passes the evidence frame to the editor. $^\dagger$ marks a closed model or planner, and Qwen-Edit is Qwen-Image-Edit.}
\label{tab:closed}
\resizebox{\linewidth}{!}{\begin{tabular}{@{}l cccc cc cc c@{}}
\toprule
System & Overall & Rec. & Res. & Reu. & Start & Stale & Fol. & Exec & $e$ shown \\
\midrule
\multicolumn{10}{@{}l}{\textit{Closed image generators}} \\
Nano Banana 2 \cite{nanobanana}$^\dagger$ & 50.4 & 57.8 & 53.3 & 40.0 & 62.7 & 29.3 & 75.4 & 88.5 & -- \\
Seedream 5 \cite{seedream5}$^\dagger$ & 45.7 & 57.8 & 46.7 & 32.6 & 60.0 & 37.3 & 68.2 & 87.5 & -- \\
Nano Banana Pro \cite{nanobananapro}$^\dagger$ & 39.3 & 42.2 & 40.4 & 35.4 & 42.7 & 29.3 & 65.8 & 89.4 & -- \\
\midrule
\multicolumn{10}{@{}l}{\textit{Agents with Qwen-Image-Edit}} \\
Gemini-3.1-Pro + Qwen-Edit$^\dagger$ & 60.2 & 62.7 & 65.3 & 52.6 & 68.0 & 70.7 & 81.2 & 88.4 & 97.8 \\
GPT-5.6 + Qwen-Edit$^\dagger$ & 55.7 & 65.8 & 54.7 & 46.6 & 62.7 & 48.0 & 77.2 & 90.0 & 97.8 \\
Kimi-K3 + Qwen-Edit$^\dagger$ & 56.7 & 59.1 & 57.8 & 53.3 & 64.0 & 60.0 & 81.2 & 87.0 & 98.2 \\
Qwen3.8-Max + Qwen-Edit$^\dagger$ & 60.9 & 64.0 & 64.0 & 54.7 & 70.7 & 66.7 & 83.0 & 88.7 & 98.2 \\
Qwen3.8 + Qwen-Edit & 19.1 & 20.4 & 14.2 & 22.6 & 24.0 & 17.3 & 36.7 & 83.3 & 60.4 \\
Kimi-K3 (no images) + Qwen-Edit$^\dagger$ & 42.2 & 46.7 & 41.3 & 38.6 & 56.0 & 42.7 & 65.0 & 86.9 & 89.4 \\
\midrule
\multicolumn{10}{@{}l}{\textit{Agents with another editor}} \\
Gemini-3.1-Pro + Qwen-Image-2.1$^\dagger$ & 55.2 & 61.4 & 62.2 & 42.0 & 62.7 & 62.7 & 83.5 & 83.8 & 97.8 \\
GPT-5.6 + GPT Image 2.5 \cite{gptimage25}$^\dagger$ & 71.0 & 75.1 & 73.3 & 64.7 & 76.0 & 78.7 & 86.7 & 92.3 & 97.8 \\
Qwen3.8 + GPT Image 2.5$^\dagger$ & 29.3 & 36.0 & 25.3 & 26.7 & 44.0 & 29.3 & 45.2 & 90.0 & 60.4 \\
\midrule
\multicolumn{10}{@{}l}{\textit{Open systems on the same pairs}} \\
\rowcolor{ours}SenseNova-U1.5 + \train & 49.4 & 46.2 & 57.3 & 44.7 & 56.0 & 62.7 & 70.1 & 87.3 & -- \\
SenseNova-U1.5 \cite{sensenova} & 29.8 & 40.4 & 31.1 & 18.0 & 45.3 & 4.0 & 47.6 & 88.5 & -- \\
FLUX.3-klein \cite{flux3klein} & 29.0 & 36.0 & 27.1 & 24.0 & 34.7 & 8.0 & 54.1 & 84.5 & -- \\
\bottomrule
\end{tabular}}
\vspace{-3mm}
\end{table}

\begin{figure}[t]
\centering
\includegraphics[width=\linewidth]{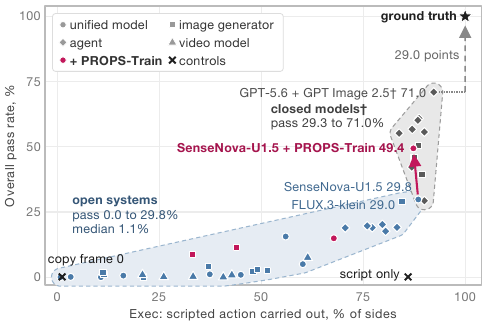}
\vspace{-6mm}
\caption{\textbf{Overall pass rate against Exec} on the 600 pairs of phase~A. Shaded regions hold the open
systems before fine-tuning and the systems with a closed model or planner, marked $^\dagger$. The dashed arrow gives the gap from the best system to the ground truth, and the solid arrow joins SenseNova-U1.5 before and after fine-tuning on \train.}
\label{fig:follow-exec}
\vspace{-3mm}
\end{figure}
\begin{figure*}[tp]
\centering
\scriptsize
\setlength{\tabcolsep}{1pt}
\renewcommand{\arraystretch}{0.6}
\includegraphics[width=\linewidth]{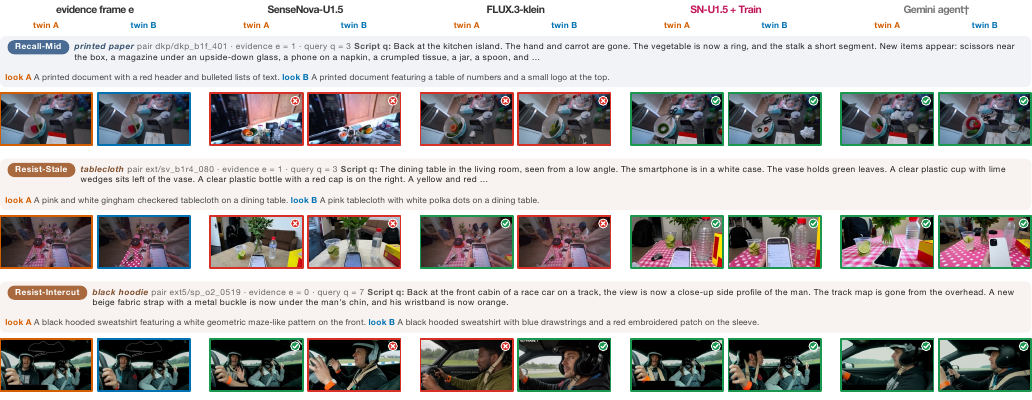}
\vspace{-2mm}
\caption{\textbf{Qualitative examples} from phase~A, chosen automatically among pairs that SenseNova-U1.5 with \train and the Gemini agent pass and that base SenseNova-U1.5 executes but fails on Follow. Each pair shows both twins, with the evidence frame outlined in the colour of its look. A green border and badge mark a side that passes and red ones a side that fails.}
\label{fig:qual}
\vspace{-3mm}
\end{figure*}

\paragraph{Carrying out the action does not mean keeping the look.} Closed image generators and the agents of \cref{tab:closed} execute the scripted action on 83 to 92\% of sides, but their Overall ranges from 19.1 to 71.0\% \crefp{fig:follow-exec}. On phase~A the highest Follow rate of an open system before fine-tuning is 55.4\%, for Emu3.5, while closed systems reach 86.7\% \crefp{app:phasea}. Given the images they accept, the open editors Qwen-Image-Edit and FireRed-Edit \cite{firered} carry out the action on only 11 to 12\% of sides, and video models do so on 21 to 61\% \crefp{app:phasea}. Failing sides mostly show neither look, and no system shows the twin's look on more than 10.7\% of sides \crefp{tab:mixbody}. Strong open models fail mostly by carrying out the action while showing neither look, SenseNova-U1.5 on 39.5\% of sides, and copying frame 0 follows on more than half of its sides but almost never carries out the change.

\begin{figure}[tp]
\centering
\includegraphics[width=\linewidth]{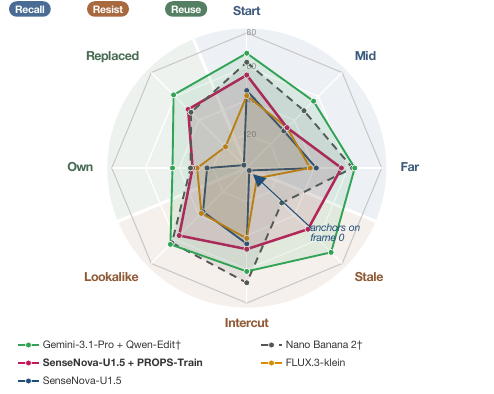}
\vspace{-6mm}
\caption{\textbf{Pair pass rate per subtype} for five systems. Open systems are scored on the full
benchmark and those marked $^\dagger$ on phase~A. \Cref{tab:mainfull,tab:phasea} give every system.}
\label{fig:subtypes}
\vspace{-3mm}
\end{figure}
\begin{table}[t]
\centering
\scriptsize
\setlength{\tabcolsep}{2pt}
\captionsetup{position=top}
\caption{\textbf{Failure composition} in \% of sides. A failing side follows its own look but misses the change, shows the twin's look in both crop orders, shows neither, or, in Reuse, loses the prop in the frame it draws at $e$, which leaves no crop to compare. Open systems are scored on the full \bench and closed systems on phase~A, and unscored sides are left out. All systems are in \cref{tab:mix}.}
\label{tab:mixbody}
\begin{adjustbox}{max width=\linewidth}\begin{tabular}{@{}l ccccccc@{}}
\toprule
 & & & \multicolumn{2}{c}{Twin's look} & \multicolumn{2}{c}{Neither look} & \\
\cmidrule(lr){4-5}\cmidrule(lr){6-7}
System & Pass & Fol., no Exec & Exec & no Exec & Exec & no Exec & Lost at $e$ \\
\midrule
\multicolumn{8}{@{}l}{\textit{Reference (full)}} \\
Copy frame 0 & 0.5 & 54.5 & 0.0 & 8.8 & 0.1 & 21.1 & 14.9 \\
Copy previous frame & 0.0 & 1.1 & 0.3 & 9.5 & 1.0 & 65.0 & 22.9 \\
\midrule
\multicolumn{8}{@{}l}{\textit{Unified multimodal models (full)}} \\
SenseNova-U1.5 \cite{sensenova} & 41.4 & 4.9 & 4.5 & 0.6 & 39.5 & 6.3 & 2.9 \\
Emu3.5 \cite{emu35} & 31.0 & 24.4 & 3.0 & 1.7 & 19.2 & 14.0 & 6.7 \\
BAGEL \cite{bagel} & 13.2 & 8.3 & 3.6 & 2.2 & 40.0 & 26.7 & 6.0 \\
\midrule
\multicolumn{8}{@{}l}{\textit{Image generators and editors (full)}} \\
FLUX.3-klein \cite{flux3klein} & 46.2 & 8.9 & 5.3 & 1.0 & 28.7 & 5.1 & 4.7 \\
Qwen-Image-2.1 \cite{qwenimage} & 6.7 & 4.7 & 2.9 & 2.1 & 36.0 & 35.6 & 11.7 \\
\midrule
\multicolumn{8}{@{}l}{\textit{Agents (open planner + editor) (full)}} \\
Qwen3.8 + Qwen-Edit & 32.5 & 5.8 & 6.3 & 1.4 & 40.7 & 8.9 & 4.3 \\
\midrule
\multicolumn{8}{@{}l}{\textit{Keyframe-conditioned video models (full)}} \\
LTX-2.5 \cite{ltxvideo} & 19.8 & 11.1 & 3.1 & 1.8 & 31.0 & 21.7 & 11.5 \\
\midrule
\multicolumn{8}{@{}l}{\textit{Fine-tuned on \train (full)}} \\
\rowcolor{ours}SenseNova-U1.5 + \train & 61.9 & 9.3 & 1.6 & 0.3 & 19.0 & 3.8 & 4.1 \\
\midrule
\multicolumn{8}{@{}l}{\textit{Closed systems and agents (phase~A)}} \\
GPT-5.6 + GPT Image 2.5 \cite{gptimage25}$^\dagger$ & 80.3 & 6.2 & 1.9 & 0.4 & 8.2 & 1.0 & 1.7 \\
Nano Banana 2 \cite{nanobanana}$^\dagger$ & 66.8 & 8.7 & 4.1 & 0.4 & 16.3 & 2.2 & 1.5 \\
\bottomrule
\end{tabular}\end{adjustbox}
\end{table}

\begin{table}[t]
\centering
\scriptsize
\setlength{\tabcolsep}{2.4pt}
\captionsetup{position=top}
\caption{\textbf{Changing the history at test time} on the 600 pairs of phase~A. The weights stay fixed and only the history changes. Removed images are replaced by grey frames, and shuffling keeps every image but not their order.}
\label{tab:counter}
\begin{tabular}{@{}l cccc cccc c@{}}
\toprule
History at test time & Overall & Rec. & Res. & Reu. & Start & Mid & Stale & Repl. & Exec \\
\midrule
\multicolumn{10}{@{}l}{\textit{SenseNova-U1.5}} \\
\quad full history & 29.8 & 40.4 & 31.1 & 18.0 & 45.3 & 33.3 & 4.0 & 8.0 & 88.5 \\
\quad frame 0 removed & 13.7 & 12.0 & 8.4 & 20.7 & 0.0 & 36.0 & 25.3 & 10.7 & 88.1 \\
\quad history shuffled & 17.2 & 25.8 & 11.1 & 14.7 & 34.7 & 32.0 & 1.3 & 0.0 & 87.3 \\
\quad previous frame only & 0.4 & 0.0 & 0.0 & 1.3 & 0.0 & 0.0 & 0.0 & 1.3 & 89.0 \\
\quad script only & 0.1 & 0.0 & 0.4 & 0.0 & 0.0 & 0.0 & 1.3 & 0.0 & 86.0 \\
\bottomrule
\end{tabular}
\vspace{-3mm}
\end{table}

\begin{figure}[tp]
\centering
\includegraphics[width=\linewidth]{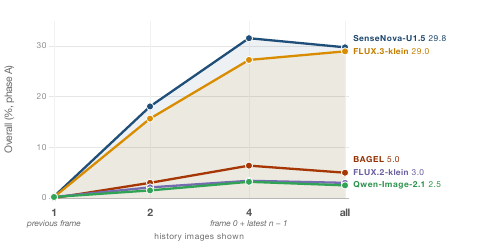}
\vspace{-6mm}
\caption{\textbf{Overall on phase~A with $n$ history images}, the previous frame for $n{=}1$ and
otherwise frame 0 and the latest $n{-}1$ frames. The script is always complete.}
\label{fig:curves}
\vspace{-3mm}
\end{figure}

\paragraph{Several generators recall a look from the opening frame far better than a later replacement.} SenseNova-U1.5 passes 45.9\% of Recall-Start pairs and still 41.2\% of Recall-Far pairs, where the gap grows to eight shots \crefp{sec:app-dist}. Yet it passes only 2.1\% of Resist-Stale pairs, whose opening frame shows a look that a later script has replaced. FLUX.3-klein falls in the same way from 42.9\% to 8.6\%, and Nano Banana~2 drops from 62.7\% to 29.3\% on phase~A \crefp{tab:closed}. To find the frames behind this, we change the history of SenseNova-U1.5 at test time with its weights fixed \crefp{tab:counter}. On phase~A, removing frame 0 raises its Resist-Stale score from 4.0\% to 25.3\%, even though Overall falls from 29.8\% to 13.7\% and Recall-Start
to zero. With only the previous frame or only the script, Overall drops to 0.4\% and 0.1\% while Exec stays at 86 to 89\%, and shuffling the history gives 17.2\%, so the order of the frames matters as well. \Cref{fig:curves} caps the number of history images and always keeps the full script. Given frame 0 and the previous frame, SenseNova-U1.5 reaches 18.1\% and keeps most of its score on the four subtypes whose evidence is frame 0, while Recall-Mid, Resist-Stale and Reuse, whose evidence lies in a later frame, drop to 0.0 to 0.7\%. With four images it already scores 31.6\%, against 29.8\% with all of them. BAGEL, Qwen-Image-2.1 and FLUX.2-klein stay at or below 6.4\% at every cap, so they accept many images but do not use them. \Cref{fig:qual} shows examples of prop drift.

\begin{figure}[t]
\centering
\includegraphics[width=\linewidth]{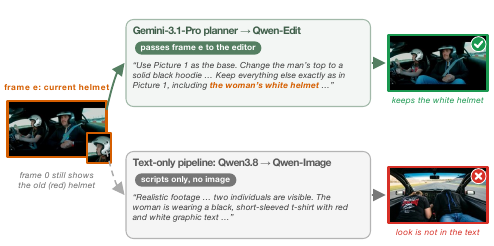}
\vspace{-6mm}
\caption{\textbf{One Resist-Stale history through two pipelines.} The closed planner passes the evidence frame $e$ to the editor and names the look to keep, while no script describes this look.}
\label{fig:agents}
\vspace{-3mm}
\end{figure}

\paragraph{A stronger planner and a stronger editor each raise the score.} Replacing the open Qwen3.8 planner with a closed planner that sees the images lifts the agent from 19.1\% to between 55.7\% and 60.9\% on phase~A \crefp{tab:closed}. These closed planners pass the evidence frame to the editor, as in \cref{fig:agents}, on 98\% of sides against 60\% for the open planner \crefp{tab:closed}. A Kimi-K3 planner that sees no images and names frames by index still passes this frame on 89\% of sides, yet its agent falls from 56.7\% to 42.2\%, and Follow on the sides given the frame from 82.9\% to 71.9\%, so seeing the images helps beyond finding the frame. Without a planner, Qwen-Image-Edit that receives the evidence frame, the previous frame and the scripts passes only 2.7\%. Keeping the plans and replacing
Qwen-Image-Edit with GPT Image 2.5 raises the GPT-5.6 agent from 55.7\% to 71.0\% and the Qwen3.8 agent from 19.1\%
to 29.3\%, with gain intervals of 11.0 to 19.8 and 6.6 to 13.6. Among open agents, replacing Qwen3.8 with Qwen3.5-9B \cite{qwen35} or Qwen-Image-Edit with FLUX.3-klein changes Overall by at most 1.1 points \crefp{app:agents}. SenseNova-U1.5 has 17.5B parameters in total, fewer than the 19.7B of the smallest of these agents, Qwen3.5-9B with FLUX.3-klein and no upsampler, and with each system given the history in its own input format it scores 11.0 points more, with an interval of 6.6 to 15.6.

\begin{table}[t]
\centering
\scriptsize
\setlength{\tabcolsep}{3pt}
\captionsetup{position=top}
\caption{\textbf{Fine-tuning on \train}, full \bench. $\Delta$ is the paired cluster-bootstrap difference to the base model before rounding, with one training run per row.}
\label{tab:sft}
\resizebox{\linewidth}{!}{\begin{tabular}{@{}l c c ccc@{}}
\toprule
Model & Overall & $\Delta$ [95\% CI] & Rec. & Res. & Reu. \\
\midrule
SenseNova-U1.5 & 26.7 & -- & 39.4 & 27.9 & 12.9 \\
\rowcolor{ours}\quad + \train & 47.0 & +20.2 [+17.1, +23.4] & 48.3 & 52.0 & 40.6 \\
\midrule
BAGEL & 4.0 & -- & 4.9 & 2.9 & 4.3 \\
\rowcolor{ours}\quad + \train & 14.9 & +10.8 [+8.6, +13.1] & 16.9 & 15.1 & 12.6 \\
\midrule
Qwen-Image-2.1 & 2.2 & -- & 2.0 & 1.9 & 2.8 \\
\rowcolor{ours}\quad + \train (LoRA) & 11.2 & +9.0 [+6.8, +11.1] & 11.2 & 13.6 & 8.9 \\
\midrule
Qwen-Image-Edit & 0.8 & -- & 1.0 & 0.2 & 1.1 \\
\rowcolor{ours}\quad + \train (LoRA) & 10.8 & +10.0 [+8.0, +12.2] & 12.8 & 12.3 & 7.2 \\
\bottomrule
\end{tabular}}
\vspace{-3mm}
\end{table}

\begin{table}[t]
\centering
\scriptsize
\setlength{\tabcolsep}{4pt}
\captionsetup{position=top}
\caption{\textbf{Removing one track} of \train, SenseNova-U1.5 on the full \bench with the same recipe for every run.}
\label{tab:ablation}
\begin{tabular}{@{}l cccc@{}}
\toprule
Training data & Overall & \textcolor{recall}{Recall} & \textcolor{resist}{Resist} & \textcolor{reuse}{Reuse} \\
\midrule
SenseNova-U1.5 & 26.7 & 39.4 & 27.9 & 12.9 \\
\rowcolor{ours}\quad + \train & 47.0 & 48.3 & 52.0 & 40.6 \\
\midrule
\quad w/o real revisits & 45.8 & 46.0 & 50.4 & 40.7 \\
\quad w/o intercut returns & 33.3 & 37.9 & 33.8 & 28.1 \\
\quad w/o synthetic chains & 44.0 & 47.6 & 47.5 & 36.8 \\
\quad w/o verified edits & 43.5 & 46.4 & 47.2 & 36.8 \\
\bottomrule
\end{tabular}
\vspace{-3mm}
\end{table}

\subsection{Training on \train}
\label{sec:sft}
\paragraph{\train reduces prop drift.} Fine-tuning raises all four models with intervals that exclude zero
\crefp{tab:sft}. SenseNova-U1.5 gains the most, 20.2 points from 26.7\% to 47.0\% on the full benchmark, and BAGEL and the two LoRA runs gain 9.0 to 10.8 points. The Follow rate of SenseNova-U1.5 rises from 46.3\% to 71.2\% while its Exec rate moves from 87.8\% to 86.1\%, so the gain comes from keeping the look. The largest gains are on updated looks, 49.2 points on Resist-Stale and 46.7 on Reuse-Replaced. On phase~A the fine-tuned model scores 49.4\%, within a point of Nano Banana~2 at 50.4\%. Three runs with different orders of the training data score 49.1\% on average with a standard deviation of 1.6 \crefp{app:stab}, and the gain stays at about 20 points when we reword the opening phrase of every phase-A script \crefp{app:para}.

\paragraph{Every track contributes.} Removing any one track of \train lowers the score of SenseNova-U1.5 \crefp{tab:ablation}. Intercut returns matter most, and without them Overall falls from 47.0\% to 33.3\%. Without real revisits Recall falls to 46.0\%, and without synthetic chains or verified edits Resist falls to 47.5 and 47.2\% and Reuse to 36.8\%. Multi-turn editing data from elsewhere helps less, as fine-tuning BAGEL on the edits of WEAVE-100k \cite{weave} raises its phase-A score from 5.0\% to 6.4\%, against 14.9\% with \train \crefp{app:general}.

\subsection{Reliability of the judge}
\label{sec:judge}\label{sec:meta}\label{app:meta}
\begin{table}[t]
\centering
\scriptsize
\setlength{\tabcolsep}{2.6pt}
\captionsetup{position=top}
\caption{\textbf{Existing consistency metrics} on 152 pairs and 7 runs, including the ground truth and the two copy baselines. Own $>$ twin is the share of sides scored higher against the own history than against the twin, with chance at 50, and for \bench the share of sides whose Follow picks the own crop. Rank places the copy of frame 0 and the ground truth among the 7 runs, using the side pass rate for \bench, and Human AUC measures how well own minus twin separates the 240 human-labelled sides by whether they follow their own look.}
\label{tab:meta}
\resizebox{\linewidth}{!}{\begin{tabular}{@{}ll ccccc@{}}
\toprule
 & & \multicolumn{2}{c}{Own $>$ twin (\%)} & \multicolumn{2}{c}{Rank of 7} & Human \\
\cmidrule(lr){3-4}\cmidrule(lr){5-6}
Metric & Compared with & GT & all runs & copy-0 & GT & AUC \\
\midrule
CLIP \cite{clip} & evidence frame $e$ & 68.4 & 60.8 & 1 & 3 & 0.59 \\
 & all history & 65.3 & 61.3 & 1 & 4 & 0.58 \\
SigLIP \cite{siglip} & evidence frame $e$ & 73.7 & 62.3 & 1 & 3 & 0.62 \\
 & all history & 69.7 & 62.3 & 1 & 3 & 0.64 \\
DINOv2 \cite{dinov2} & evidence frame $e$ & 63.2 & 59.8 & 1 & 3 & 0.72 \\
 & all history & 59.9 & 60.2 & 1 & 4 & 0.66 \\
DreamSim \cite{dreamsim} & evidence frame $e$ & 85.9 & 66.8 & 1 & 3 & 0.74 \\
 & all history & 79.6 & 67.5 & 1 & 4 & 0.70 \\
\midrule
VLM rating & Gemma-4-31B, 0--10 & 68.4 & 57.7 & 6 & 2 & 0.66 \\
VLM rating & Qwen3.8-27B, 0--10 & 57.4 & 52.0 & 7 & 4 & 0.55 \\
VLM rating & GPT-5.5, 0--10 & 76.0 & 60.9 & 6 & 3 & 0.64 \\
Trajectory rating & Gemma-4-31B, 1--5 & 68.3 & 58.3 & 6 & 1 & 0.67 \\
Trajectory rating & Qwen3.8-27B, 1--5 & 61.3 & 55.1 & 6 & 3 & 0.53 \\
\midrule
\bench (ours) & twin crops & 100.0 & 71.2 & 6 & 1 & 0.85 \\
\bottomrule
\end{tabular}}
\vspace{-3mm}
\end{table}

\paragraph{Similarity metrics reward copying and barely separate the twins.} A metric of prop consistency should score an output higher against its own history than against the twin history, and rank the ground truth above a copied frame. We test this on 152 pairs, 19 per subtype, with the outputs of seven runs that include the ground truth and both copy baselines \crefp{tab:meta}. Embedding similarities from CLIP \cite{clip}, SigLIP \cite{siglip}, DINOv2 \cite{dinov2} and DreamSim \cite{dreamsim} prefer the own history on only 60 to 68\% of sides, even when compared with the evidence frame, and all four rank the copy of frame 0 first of the seven runs, above the ground truth. VLM ratings on a 0 to 10 scale prefer the own history on 52 to 61\% of sides, and rating the whole trajectory, as LongMIG-Bench \cite{longmig} does, on 55.1 to 58.3\%. These ratings are also unstable, since Qwen3.8 and GPT-5.5 give the same score twice in at most two thirds of the cases. Against human labels the twin crops of \bench separate the sides that keep their own look with an AUC of 0.85, against 0.53 to 0.74 for the other metrics \crefp{tab:meta,fig:human}, and even DreamSim on a crop of $X$ at the ground-truth box reaches only 0.74.

\paragraph{The judge agrees with human annotation.} A human annotator blind to the judge's answers labels 240 judge decisions, sampled across subtypes, five systems and judge outcomes, and agrees with the judge on 86.2\% of side outcomes, with Cohen's $\kappa$ of 0.68. On the 77 decisions that a second annotator also labelled, the two annotators agree on 87.0\%, and the judge agrees with each of them on 90.9 and 85.7\%. The pass rate is 31.2\% by the judge and 30.0\% by human annotation, and the rankings of the five systems correlate with a Spearman $\rho$ of 0.80 \crefp{fig:human}. Since the same Gemma-4 prefers the matching history on only 57.7\% of sides when it rates whole histories \crefp{tab:meta}, the agreement comes from asking it to choose between two crops.

\paragraph{A second judge gives a similar ranking.} Qwen3.8-27B, from another model family, ranks the 16 systems scored by both judges, eight closed and eight open, with Kendall $\tau=0.97$ \crefp{app:stab}. Nano Banana~2, a Google model like Gemma-4, scores 50.4\% under both judges. SenseNova-U1.5 + \train moves the most, from 49.4\% under Gemma-4 to 44.3\% under Qwen3.8, so its gain over the base falls from 19.6 to 14.8 points.

\section{Conclusion}
\label{sec:conclusion}
On the twin histories of \bench, all 34 systems we test show prop drift. Several generators keep an opening look far better than its later replacement, and swapping the parts of agents shows that both history understanding and image generation limit agent performance. Fine-tuning on \train clearly improves all four tested models.
Twin histories can also test visual instructions, reasoning with generated images and world models.

{\small
\bibliographystyle{ieeenat_fullname}
\bibliography{refs}
}

\clearpage

\appendix
\makeatletter
\setcounter{topnumber}{4}\setcounter{bottomnumber}{2}\setcounter{totalnumber}{6}\setcounter{dbltopnumber}{3}
\renewcommand{\topfraction}{0.95}\renewcommand{\bottomfraction}{0.6}\renewcommand{\dbltopfraction}{0.95}
\renewcommand{\textfraction}{0.05}\renewcommand{\floatpagefraction}{0.8}\renewcommand{\dblfloatpagefraction}{0.8}
\setlength{\@fptop}{0pt}\setlength{\@dblfptop}{0pt}
\makeatother

\section{Benchmark construction details}
\label{app:build}
\begin{figure*}[t]
\centering
\includegraphics[width=\linewidth]{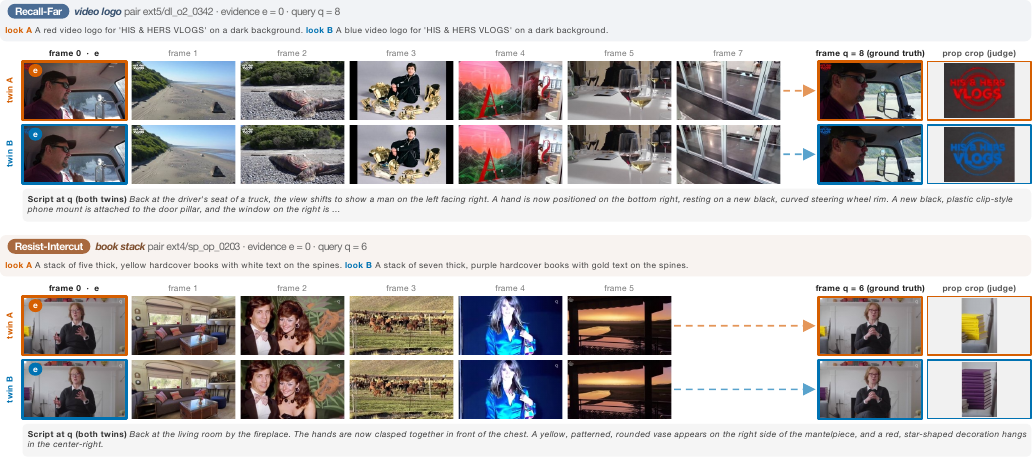}
\vspace{-5mm}
\caption{\textbf{Two \bench pairs.} The twins share every script, and the prop takes the look of
\textcolor{lookA}{twin A} or \textcolor{lookB}{twin B}. The evidence frame $e$ is outlined and the judge's reference
crops are on the right. Top, Recall-Far, with frame 6 left out for space: the logo was last seen in frame 0, eight shots earlier. Bottom,
Resist-Intercut: the chain alternates with another storyline until $q$.}
\label{fig:examples}
\vspace{-3mm}
\end{figure*}

\begin{figure*}[t]
\centering
\includegraphics[width=0.665\linewidth]{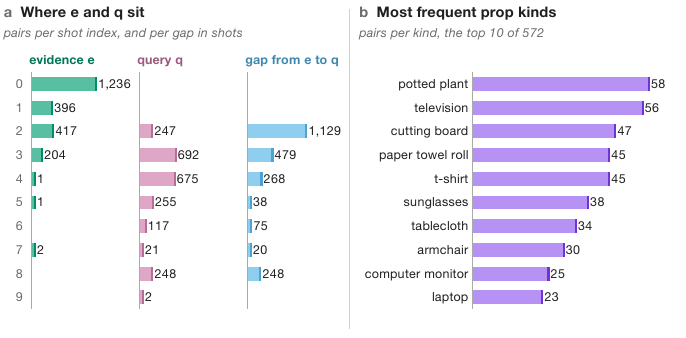}
\vspace{-2mm}
\caption{\textbf{\bench statistics.} \textbf{a} Pairs per shot index of the evidence $e$ and of the query $q$, and
per gap between them. \textbf{b} The ten most frequent of the 572 prop kinds.}
\label{fig:stats}
\vspace{-2mm}
\end{figure*}

This section follows the four steps of \cref{fig:pipeline}. \Cref{fig:viz-build} shows them on one pair per
subtype, and \cref{fig:examples} shows two finished pairs as stored.

\paragraph{Step 1, select a triple.} A VLM grounds the prop $X$ in $e$ and $q$, names its kind and its current look,
and proposes alternative looks for the twins. A triple is kept only if $X$ is visible in both frames,
no frame of the same place in between shows it, and no script up to $q$ describes it.

\paragraph{Step 2, edit $e$ and $q$ together.} One edit call changes $X$ in both crops into look A, and another changes it into look B. In a pilot, the judge accepted 30 of 40 pairs edited this way and 20 to 25 of 40 pairs whose
two frames were edited one at a time. The edited region is pasted back into the source frames inside an extended
SAM\,2.1 mask, and we call the share of the changed pixels that the paste-back loses the edit leak. We keep the best of three seeds. In Recall and Resist, where the edited frames enter the history, we require an edit leak of at most one half. In Reuse the model draws $e$ and $q$ itself, so the edited frames serve only as the true frames of the final filter. Both looks go through the same edit and
paste-back.

\paragraph{Step 3, check both twins.} Besides the conditions in \cref{sec:construction}, every execution question
must be answerable from the history and the scripts. Two VLMs check this, and when they do not both confirm it, a sentence is added to the script. After these checks a query keeps one to nine execution questions, 2.2 on average.

\paragraph{Step 4, compose and filter.} The evidence is frame 0 in Recall-Start, Recall-Far, Resist-Intercut and
Resist-Lookalike, 1,236 pairs in all, and a later frame in the other four subtypes. Recall-Far inserts frames of the
same size from other chains, none showing an object of the kind of $X$. In Resist-Stale the story returns to the
place two steps before $q$, the script says that $X$ has been replaced, and the frame shows the current look.
Resist-Intercut alternates the two chains, $a_0\,b_0\,a_1\,b_1\cdots a_q$, and the other storyline preferably contains
an object of the same kind. In Resist-Lookalike the inserted frame comes from another chain, with an object of the
same kind edited into the other twin's look. In Reuse-Replaced the opening frame is not edited, so both twins share
it. In both Reuse subtypes the generator draws only $e$ and $q$, and every other history frame is the ground-truth
frame. The final filter scores the true frames of each composed pair with the main judge, Gemma-4-31B.

\paragraph{The textual route in Reuse.} Reuse leaves a textual route open on purpose. The look is named once, in the
script of the step the model draws itself, as a shot script names a newly introduced prop, and never again. Given only the scripts, SenseNova-U1.5 passes 0.0\% of Reuse pairs \crefp{tab:counter}. A
text-only pipeline, in which Qwen3.8 turns the scripts into a prompt for Qwen-Image-2512, passes 42.6\% of the Reuse pairs
of phase~A, 0.4\% of Recall and 0.0\% of Resist, so the textual route helps in Reuse but solves no requirement. If a
model draws the same look at $e$ on both sides, its two crops look alike, and since the question is asked in both
orders, the pair passes only if the judge still tells them apart.

\section{Evaluated systems}
\label{app:models}
\begin{table}[t]
\centering
\scriptsize
\setlength{\tabcolsep}{2.4pt}
\captionsetup{position=top}
\caption{\textbf{Evaluated checkpoints.} Open weights are Hugging Face repository names, and closed systems are API model names.}
\label{tab:models}
\begin{tabular}{@{}ll@{}}
\toprule
System & Checkpoint / API model \\
\midrule
\multicolumn{2}{@{}l}{\textit{Unified multimodal models}} \\
SenseNova-U1.5 & sensenova/SenseNova-U1.5-8B-MoT \\
Emu3.5 & BAAI/Emu3.5 (+ Emu3.5-VisionTokenizer) \\
BAGEL & ByteDance-Seed/BAGEL-7B-MoT \\
HiDream-O1 & HiDream-ai/HiDream-O1-Image \\
LLaDA2-Uni & inclusionAI/LLaDA2.0-Uni \\
OmniGen2 & OmniGen2/OmniGen2 \\
Lumina-DiMOO & Alpha-VLLM/Lumina-DiMOO \\
Janus-Pro & deepseek-ai/Janus-Pro-7B \\
Show-o2 & showlab/show-o2-7B \\
\midrule
\multicolumn{2}{@{}l}{\textit{Image generators and editors}} \\
FLUX.3-klein & black-forest-labs/flux-3-image-klein \\
FLUX.2-klein & black-forest-labs/FLUX.2-klein-9B \\
VINCIE & ByteDance-Seed/VINCIE-7B \\
Qwen-Image-2.1 & Qwen/Qwen-Image-2.1 \\
FLUX.2-dev & black-forest-labs/FLUX.2-dev \\
FireRed-Edit & FireRedTeam/FireRed-Image-Edit-1.1 \\
Qwen-Image-Edit & Qwen/Qwen-Image-Edit-2511 \\
FLUX.1-Kontext & black-forest-labs/FLUX.1-Kontext-dev \\
\midrule
\multicolumn{2}{@{}l}{\textit{Agents (planner + editor)}} \\
Open planners & Qwen3.8-27B, Qwen3.5-9B \\
Closed planners & gemini-3.1-pro-preview, gpt-5.6-sol, kimi-k3, qwen3.8-max \\
Open editors & Qwen-Image-Edit-2511, Qwen-Image-2.1, flux-3-image-klein \\
Text-only pipeline & Qwen3.8-27B with Qwen/Qwen-Image-2512 \\
Closed editor & gpt-image-2.5-sunburst \\
\midrule
\multicolumn{2}{@{}l}{\textit{Closed image generators}} \\
Nano Banana 2 / Pro & gemini-3.1-flash-image / gemini-3-pro-image \\
Seedream 5 & doubao-seedream-5-0-260128 \\
\midrule
\multicolumn{2}{@{}l}{\textit{Keyframe-conditioned video models}} \\
LTX-2.5 & ltx-2.5-22b-dev (+ distilled LoRA, $2\times$ upscaler) \\
Wan-VACE & Wan-AI/Wan2.1-VACE-14B-diffusers \\
StoryMem & Kevin-thu/StoryMem (Wan2.2-MI2V-A14B) \\
HunyuanVideo-I2V & HunyuanVideo-1.5-Diffusers-480p\_i2v \\
Wan-TI2V & Wan-AI/Wan2.2-TI2V-5B-Diffusers \\
Wan-I2V & Wan-AI/Wan2.2-I2V-A14B-Diffusers \\
\midrule
\multicolumn{2}{@{}l}{\textit{Judges}} \\
Main / second judge & google/gemma-4-31B-it / Qwen3.8-27B \\
Rating judge, \cref{sec:meta} & gpt-5.5 \\
\bottomrule
\end{tabular}

\vspace{-3mm}
\end{table}

\begin{table*}[tp]
\centering
\scriptsize
\setlength{\tabcolsep}{3.2pt}
\captionsetup{position=top}
\caption{\textbf{All open systems and the reference rows on the full \bench}, with the columns of \cref{tab:main}.}
\label{tab:mainfull}
\begin{tabular}{@{}l c ccc ccc ccc cc cc@{}}
\toprule
 & & \multicolumn{3}{c}{Groups} & \multicolumn{3}{c}{\textcolor{recall}{Recall}} & \multicolumn{3}{c}{\textcolor{resist}{Resist}} & \multicolumn{2}{c}{\textcolor{reuse}{Reuse}} & \multicolumn{2}{c}{Diagnostic} \\
\cmidrule(lr){3-5}\cmidrule(lr){6-8}\cmidrule(lr){9-11}\cmidrule(lr){12-13}\cmidrule(l){14-15}
System & Overall & Recall & Resist & Reuse & Start & Mid & Far & Stale & Intercut & Lookalike & Own & Replaced & Follow & Exec \\
\midrule
\multicolumn{15}{@{}l}{\textit{Reference}} \\
Ground-truth frames & 100.0 & 100.0 & 100.0 & 100.0 & 100.0 & 100.0 & 100.0 & 100.0 & 100.0 & 100.0 & 100.0 & 100.0 & 100.0 & 100.0 \\
Copy frame 0 & 0.1 & 0.1 & 0.2 & 0.0 & 0.2 & 0.0 & 0.0 & 0.0 & 0.5 & 0.0 & 0.0 & 0.0 & 54.9 & 0.7 \\
Copy previous frame & 0.0 & 0.0 & 0.0 & 0.0 & 0.0 & 0.0 & 0.0 & 0.0 & 0.0 & 0.0 & 0.0 & 0.0 & 1.1 & 1.7 \\
\midrule
\multicolumn{15}{@{}l}{\textit{Unified multimodal models}} \\
SenseNova-U1.5 \cite{sensenova} & \textbf{26.7} & \textbf{39.4} & \textbf{27.9} & \textbf{12.9} & \textbf{45.9} & \textbf{31.1} & \textbf{41.2} & 2.1 & \textbf{44.9} & \textbf{36.6} & \textbf{23.5} & 2.4 & 46.3 & \textbf{87.8} \\
Emu3.5 \cite{emu35} & 15.9 & 19.6 & 15.7 & 12.3 & 24.9 & 19.1 & 14.9 & \textbf{9.6} & 20.9 & 16.6 & 12.3 & \textbf{12.3} & \textbf{55.4} & 57.0 \\
BAGEL \cite{bagel} & 4.0 & 4.9 & 2.9 & 4.3 & 7.2 & 5.3 & 2.3 & 2.1 & 2.7 & 3.9 & 2.7 & 5.9 & 21.5 & 60.5 \\
HiDream-O1 \cite{hidreamo1} & 1.6 & 1.8 & 0.9 & 2.1 & 3.5 & 1.8 & 0.0 & 0.0 & 1.6 & 1.0 & 3.7 & 0.5 & 10.5 & 39.7 \\
LLaDA2-Uni \cite{llada2uni} & 1.4 & 1.5 & 0.9 & 1.8 & 3.5 & 0.9 & 0.0 & 2.1 & 0.0 & 0.5 & 1.6 & 1.9 & 11.3 & 49.2 \\
OmniGen2 \cite{omnigen2} & 0.7 & 0.3 & 1.2 & 0.5 & 0.5 & 0.4 & 0.0 & 3.2 & 0.0 & 0.5 & 0.5 & 0.5 & 13.6 & 17.1 \\
Lumina-DiMOO \cite{lumina} & 0.1 & 0.2 & 0.2 & 0.0 & 0.5 & 0.0 & 0.0 & 0.0 & 0.5 & 0.0 & 0.0 & 0.0 & 14.1 & 2.9 \\
Janus-Pro \cite{januspro} & 0.0 & 0.0 & 0.0 & 0.0 & 0.0 & 0.0 & 0.0 & 0.0 & 0.0 & 0.0 & 0.0 & 0.0 & 1.0 & 1.4 \\
Show-o2 \cite{showo2} & 0.0 & 0.0 & 0.0 & 0.0 & 0.0 & 0.0 & 0.0 & 0.0 & 0.0 & 0.0 & 0.0 & 0.0 & 1.4 & 11.9 \\
\midrule
\multicolumn{15}{@{}l}{\textit{Image generators and editors}} \\
FLUX.3-klein \cite{flux3klein} & \textbf{30.4} & \textbf{38.1} & \textbf{29.4} & \textbf{23.6} & \textbf{42.9} & \textbf{33.8} & \textbf{37.6} & \textbf{8.6} & \textbf{41.7} & \textbf{38.0} & \textbf{29.4} & \textbf{17.8} & \textbf{55.1} & \textbf{84.2} \\
FLUX.2-klein \cite{flux2} & 3.1 & 2.5 & 1.7 & 5.1 & 4.5 & 3.1 & 0.0 & 1.1 & 1.1 & 2.9 & 5.9 & 4.3 & 16.3 & 49.4 \\
VINCIE \cite{vincie} & 2.9 & 2.9 & 4.0 & 1.8 & 6.9 & 1.3 & 0.5 & 3.7 & 4.8 & 3.4 & 0.5 & 3.1 & 18.9 & 40.8 \\
Qwen-Image-2.1 \cite{qwenimage} & 2.2 & 2.0 & 1.9 & 2.8 & 3.2 & 2.7 & 0.0 & 0.5 & 3.7 & 1.5 & 4.3 & 1.4 & 11.4 & 51.8 \\
FLUX.2-dev \cite{flux2} & 2.0 & 1.6 & 1.4 & 2.9 & 2.6 & 2.2 & 0.0 & 1.1 & 1.6 & 1.5 & 4.8 & 0.9 & 10.5 & 49.9 \\
FireRed-Edit \cite{firered} & 1.5 & 2.2 & 0.3 & 2.1 & 4.2 & 1.8 & 0.5 & 0.5 & 0.0 & 0.5 & 0.5 & 3.8 & 32.8 & 16.0 \\
Qwen-Image-Edit \cite{qwenimage} & 0.8 & 1.0 & 0.2 & 1.1 & 2.2 & 0.9 & 0.0 & 0.5 & 0.0 & 0.0 & 1.6 & 0.5 & 18.9 & 16.6 \\
FLUX.1-Kontext \cite{fluxkontext} & 0.0 & 0.0 & 0.0 & 0.0 & 0.0 & 0.0 & 0.0 & 0.0 & 0.0 & 0.0 & 0.0 & 0.0 & 2.5 & 21.9 \\
\midrule
\multicolumn{15}{@{}l}{\textit{Agents (open planner + editor)}} \\
Qwen3.8 + Qwen-Edit & \textbf{19.6} & \textbf{20.6} & \textbf{14.1} & \textbf{24.2} & \textbf{22.5} & 27.6 & \textbf{11.8} & \textbf{18.2} & \textbf{9.6} & \textbf{14.6} & 31.0 & \textbf{17.5} & \textbf{38.4} & \textbf{83.1} \\
Qwen3.8 + Qwen-Image-2.1 & 16.9 & 17.0 & 10.8 & 23.0 & 13.2 & \textbf{30.7} & 7.2 & 17.6 & 5.9 & 8.8 & \textbf{32.6} & 13.3 & 35.2 & 79.6 \\
\midrule
\multicolumn{15}{@{}l}{\textit{Keyframe-conditioned video models}} \\
LTX-2.5 \cite{ltxvideo} & \textbf{8.3} & \textbf{12.1} & \textbf{8.9} & \textbf{4.0} & \textbf{15.4} & \textbf{13.3} & \textbf{7.7} & \textbf{5.9} & \textbf{10.2} & \textbf{10.7} & \textbf{4.3} & \textbf{3.8} & \textbf{30.9} & \textbf{60.8} \\
Wan-VACE \cite{vace} & 0.8 & 1.4 & 0.5 & 0.5 & 2.1 & 2.2 & 0.0 & 1.6 & 0.0 & 0.0 & 0.5 & 0.5 & 11.6 & 24.0 \\
StoryMem \cite{storymem} & 0.4 & 0.2 & 0.5 & 0.5 & 0.6 & 0.0 & 0.0 & 1.6 & 0.0 & 0.0 & 0.0 & 0.9 & 7.5 & 46.4 \\
HunyuanVideo-I2V \cite{hunyuanvideo} & 0.1 & 0.1 & 0.0 & 0.3 & 0.3 & 0.0 & 0.0 & 0.0 & 0.0 & 0.0 & 0.5 & 0.2 & 4.7 & 34.3 \\
Wan-TI2V \cite{wan} & 0.0 & 0.1 & 0.0 & 0.0 & 0.0 & 0.4 & 0.0 & 0.0 & 0.0 & 0.0 & 0.0 & 0.0 & 2.1 & 30.1 \\
Wan-I2V \cite{wan} & 0.0 & 0.0 & 0.0 & 0.1 & 0.0 & 0.0 & 0.0 & 0.0 & 0.0 & 0.0 & 0.0 & 0.2 & 3.2 & 43.9 \\
\midrule
\multicolumn{15}{@{}l}{\textit{Fine-tuned on \train (ours)}} \\
\rowcolor{ours}SenseNova-U1.5 + \train & \textbf{47.0} & \textbf{48.3} & \textbf{52.0} & \textbf{40.6} & \textbf{55.1} & \textbf{33.8} & \textbf{56.1} & \textbf{51.3} & \textbf{48.1} & \textbf{56.6} & \textbf{32.1} & \textbf{49.1} & \textbf{71.2} & \textbf{86.1} \\
\rowcolor{ours}BAGEL + \train & 14.9 & 16.9 & 15.1 & 12.6 & 25.0 & 11.1 & 14.5 & 11.2 & 16.6 & 17.6 & 9.1 & 16.1 & 41.9 & 69.7 \\
\rowcolor{ours}Qwen-Image-2.1 + \train (LoRA) & 11.2 & 11.2 & 13.6 & 8.9 & 20.5 & 6.7 & 6.3 & 15.0 & 10.2 & 15.6 & 4.3 & 13.5 & 50.1 & 45.7 \\
\rowcolor{ours}Qwen-Image-Edit + \train (LoRA) & 10.8 & 12.8 & 12.3 & 7.2 & 25.0 & 7.6 & 5.9 & 8.6 & 16.0 & 12.2 & 3.7 & 10.7 & 52.4 & 38.7 \\
\bottomrule
\end{tabular}
\vspace{-3mm}
\end{table*}

\begin{table}[t]
\centering
\scriptsize
\setlength{\tabcolsep}{3pt}
\captionsetup{position=top}
\caption{\textbf{Models with a capped input.} Top: models that take at most three images, on the pairs whose evidence frame they receive, with SenseNova-U1.5 and the agent on the same pairs. Bottom: models that take one image.}
\label{tab:full}
\begin{tabular}{@{}l cccc cc@{}}
\toprule
System & Overall & Rec. & Res. & Reu. & Follow & Exec \\
\midrule
\multicolumn{7}{@{}l}{\textit{Pairs whose evidence is within 3 images (2,118 pairs)}} \\
SenseNova-U1.5 \cite{sensenova} & 26.6 & 39.5 & 27.9 & 12.3 & 46.3 & 87.8 \\
Qwen3.8 + Qwen-Edit & 19.6 & 21.2 & 14.1 & 23.4 & 37.8 & 82.9 \\
FireRed-Edit \cite{firered} & 1.7 & 2.5 & 0.3 & 2.3 & 34.9 & 16.6 \\
Qwen-Image-Edit \cite{qwenimage} & 0.9 & 1.2 & 0.2 & 1.4 & 20.0 & 17.3 \\
OmniGen2 \cite{omnigen2} & 0.8 & 0.4 & 1.2 & 0.7 & 14.4 & 17.3 \\
\midrule
\multicolumn{7}{@{}l}{\textit{Single-image input models (latest frame + script)}} \\
Show-o2 \cite{showo2} & 0.0 & 0.0 & 0.0 & 0.0 & 1.4 & 11.9 \\
FLUX.1-Kontext \cite{fluxkontext} & 0.0 & 0.0 & 0.0 & 0.0 & 2.5 & 21.9 \\
Wan-I2V \cite{wan} & 0.0 & 0.0 & 0.0 & 0.1 & 3.2 & 43.9 \\
Wan-TI2V \cite{wan} & 0.0 & 0.1 & 0.0 & 0.0 & 2.1 & 30.1 \\
HunyuanVideo-I2V \cite{hunyuanvideo} & 0.1 & 0.1 & 0.0 & 0.3 & 4.7 & 34.3 \\
\bottomrule
\end{tabular}
\vspace{-3mm}
\end{table}

\paragraph{Input format.} Every system receives the interleaved history in its native format, up to the largest
number of images it was trained with or its context accepts. Show-o2 takes one image, Lumina-DiMOO two, its trained maximum with a 4,096-token context, and FireRed-Edit, Qwen-Image-Edit and OmniGen2 three. A system that takes fewer images than the history holds receives frame 0 and the latest frames, or only the previous frame if it takes one. Wan-VACE takes the previous keyframe as its first frame and up to three earlier keyframes as references, and LLaDA2-Uni drops the oldest images after frame 0 until the sequence fits 8,192 positions. Janus-Pro, the other UMMs and the three
closed image generators receive all images. Keyframe-conditioned video models generate one clip per shot from the previous keyframe, the history images
they accept as references and the shot script, and we score the last frame of the clip. Image editors receive the
previous frame and the accumulated script, and multi-reference editors also receive the earlier frames as
references. The agent planner sees the full script and every history image. It selects at most three frames and writes one edit instruction, and the editor receives only these
frames and that instruction. Every history frame is a ground-truth keyframe, except that in Reuse the frame a model draws at $e$ stays in its history. Resolution, sampler and step count follow each model's official recommendation, and
model versions are listed in \cref{tab:models}.

\paragraph{Models limited to a few images.} Systems that accept only two or three images cannot see the evidence
of every pair. Besides their scores on all pairs \crefp{tab:mainfull}, we report the models that take three images on the pairs whose evidence frame is among the images they receive \crefp{tab:full}. These also stay below 2\%, so their low numbers do not come from the cap.

\section{Stability}
\label{app:stab}
\begin{table}[t]
\centering
\scriptsize
\setlength{\tabcolsep}{2.6pt}
\captionsetup{position=top}
\caption{\textbf{Stability} of Overall on phase~A. Seeds change only the generation seed, and for SenseNova-U1.5 + \train they change the order of the training data. Re-sampled scores the same images again with another judge sampling salt, and Qwen3.8 scores them with Qwen3.8-27B instead of Gemma-4, and -- marks a run it did not score. For the other systems the fourth column gives the Gemma-4 score.}
\label{tab:stability}
\begin{adjustbox}{max width=\linewidth}\begin{tabular}{@{}l ccc c c c@{}}
\toprule
 & \multicolumn{4}{c}{Seed} & \multicolumn{2}{c}{Judge} \\
\cmidrule(lr){2-5}\cmidrule(l){6-7}
System & 0 & 1 & 2 & mean $\pm$ SD & re-sampled & Qwen3.8 \\
\midrule
SenseNova-U1.5 & 29.8 & 29.5 & 27.4 & 28.9 $\pm$ 1.3 & -- & 29.5 \\
SenseNova-U1.5 + \train & 49.4 & 47.4 & 50.5 & 49.1 $\pm$ 1.6 & -- & 44.3 \\
FLUX.3-klein & 29.0 & 31.6 & 29.5 & 30.0 $\pm$ 1.4 & -- & 29.9 \\
Emu3.5 & 15.6 & 15.3 & 15.1 & 15.3 $\pm$ 0.3 & -- & -- \\
BAGEL & 5.0 & 3.4 & 3.4 & 3.9 $\pm$ 0.9 & -- & 2.8 \\
LTX-2.5 & 7.5 & 7.2 & 8.6 & 7.8 $\pm$ 0.7 & -- & 7.7 \\
Qwen-Image-2.1 & 2.5 & 2.9 & 3.0 & 2.8 $\pm$ 0.3 & -- & -- \\
Qwen-Image-Edit & 1.0 & -- & -- & -- & 1.0 & 1.1 \\
FLUX.2-klein & 3.0 & -- & -- & -- & 3.0 & -- \\
\midrule
\multicolumn{7}{@{}l}{\textit{Other systems scored by both judges, Gemma-4 / Qwen3.8}} \\
Qwen3.8-Max + Qwen-Edit$^\dagger$ & -- & -- & -- & 60.9 & -- & 59.3 \\
Gemini-3.1-Pro + Qwen-Edit$^\dagger$ & -- & -- & -- & 60.2 & -- & 57.9 \\
Kimi-K3 + Qwen-Edit$^\dagger$ & -- & -- & -- & 56.7 & -- & 56.4 \\
GPT-5.6 + Qwen-Edit$^\dagger$ & -- & -- & -- & 55.7 & -- & 52.9 \\
Nano Banana 2 \cite{nanobanana}$^\dagger$ & -- & -- & -- & 50.4 & -- & 50.4 \\
Seedream 5 \cite{seedream5}$^\dagger$ & -- & -- & -- & 45.7 & -- & 43.2 \\
Kimi-K3 (no images) + Qwen-Edit$^\dagger$ & -- & -- & -- & 42.2 & -- & 41.4 \\
Nano Banana Pro \cite{nanobananapro}$^\dagger$ & -- & -- & -- & 39.3 & -- & 40.1 \\
Wan-VACE \cite{vace} & -- & -- & -- & 0.9 & -- & 0.7 \\
OmniGen2 \cite{omnigen2} & -- & -- & -- & 0.6 & -- & 1.0 \\
\bottomrule
\end{tabular}\end{adjustbox}
\vspace{-3mm}
\end{table}

\cref{tab:stability} repeats generation with other seeds and re-runs the judge.
Over three generation seeds SenseNova-U1.5 scores 28.9 $\pm$ 1.3 on phase~A. Fine-tuned three times on \train with different orders of the training data, it scores 49.4, 47.4 and 50.5\%, or 49.1 $\pm$ 1.6, so the gain of about 20 points is far outside both spreads. The other systems have standard
deviations of at most 1.4 points, with FLUX.3-klein at 30.0 $\pm$ 1.4, Emu3.5 15.3 $\pm$ 0.3, LTX-2.5 7.8 $\pm$ 0.7, BAGEL
3.9 $\pm$ 0.9 and Qwen-Image-2.1 2.8 $\pm$ 0.3, so seeds move scores far less than the gaps between families. Only
FLUX.3-klein and SenseNova-U1.5 overlap on phase~A, and on the full benchmark FLUX.3-klein leads by 3.6 points with an interval of 0.3 to 6.9. Re-scoring the same images with a different judge sampling salt leaves Qwen-Image-Edit and FLUX.2-klein unchanged. A second judge, Qwen3.8-27B, scores 16 systems on phase~A. The eight closed ones are the three closed image generators and the five agents with a closed planner and Qwen-Image-Edit, the planner without images included. The eight open ones are SenseNova-U1.5 before and after fine-tuning, FLUX.3-klein, Qwen-Image-Edit, BAGEL, LTX-2.5, OmniGen2 and Wan-VACE. \Cref{tab:stability} lists the scores of all 16 under both judges. Qwen3.8 keeps the ranking of the 16 systems with Kendall $\tau = 0.97$ and moves each of them by at most 5.1 points, the most for SenseNova-U1.5 + \train, from 49.4 to 44.3\%. Of the 18 pairs of these systems within 5 points of each other under Gemma-4, 16 keep their order. 

\section{Context length}
\label{app:ctx}
\Cref{fig:curves} in \cref{sec:counter} shows Overall on phase~A when the number of history images is capped, with the
full script always given. With only the previous frame in the history every model passes 0.0 to 0.3\% of pairs, and SenseNova-U1.5 passes 0.4\% when the other frames are replaced by grey ones \crefp{tab:counter}. With frame 0 and
the previous frame, SenseNova-U1.5 reaches 18.1\%. On the four subtypes whose evidence is frame 0 it keeps most of
its score, with 48.0\% on Recall-Start, 37.3\% on Recall-Far, 40.0\% on Resist-Intercut and 36.0\% on
Resist-Lookalike, against 45.3, 42.7, 52.0 and 37.3\% with the full history. Recall-Mid, Resist-Stale and Reuse,
whose evidence lies in a later frame, drop to 0.0 to 0.7\%. Without frame 0, SenseNova-U1.5 passes none of the phase-A pairs of those four subtypes. With four images it already scores 31.6\%,
against 29.8\% with all of them. BAGEL, Qwen-Image-2.1 and FLUX.2-klein stay at or below 6.4\% at every cap, so they accept
many images but do not use them. FLUX.3-klein behaves like SenseNova-U1.5, with 0.2\% given the previous frame,
15.7\% given frame 0 and the previous frame, and 27.3\% given four images, against 29.0\% with all of them.

\section{Judge protocol}
\label{app:judge}
\begin{figure*}[t]
\centering
\includegraphics[width=\linewidth]{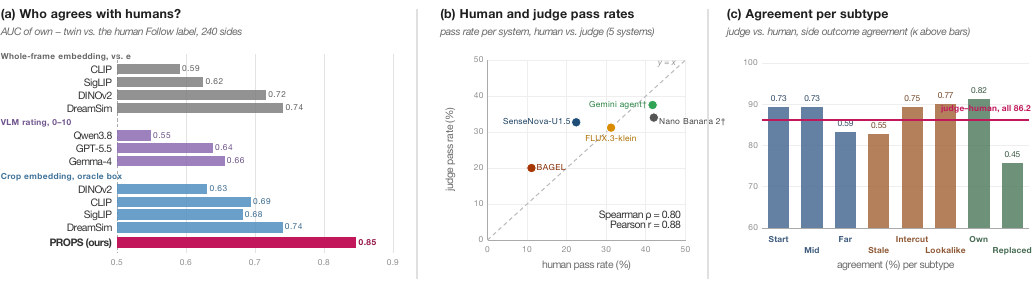}
\vspace{-5mm}
\caption{\textbf{Agreement with human labels} on 240 sides of five systems. \textbf{a} AUC of the own minus
twin score of each metric at predicting the human Follow label. \textbf{b} Pass rate of each system by human and
judge. \textbf{c} Agreement of judge and human on the side outcome per subtype, with $\kappa$ above the bars.}
\label{fig:human}
\vspace{-3mm}
\end{figure*}

\begin{table*}[tp]
\centering
\scriptsize
\setlength{\tabcolsep}{4pt}
\captionsetup{position=top}
\caption{\textbf{Human annotation against the judge} per subtype and system, 240 labelled sides, with $\kappa$ after each agreement. Every row has between 28 and 50 sides.}
\label{tab:human}
\begin{adjustbox}{max width=\linewidth}\begin{tabular}{@{}l c cc c c c c@{}}
\toprule
 & Sides & Crop choice & Follow & Questions & Exec questions & Side pass & Pass rate \\
 & & \% ($\kappa$) & \% ($\kappa$) & & \% ($\kappa$) & \% ($\kappa$) & human / judge \\
\midrule
All & 240 & 66.7 (0.50) & 86.2 (0.70) & 530 & 90.6 (0.56) & 86.2 (0.68) & 30.0 / 31.2 \\
\midrule
Recall-Start & 28 & 64.3 (0.46) & 89.3 (0.77) & 68 & 92.6 (0.72) & 89.3 (0.73) & -- \\
Recall-Mid & 28 & 53.6 (0.31) & 82.1 (0.58) & 65 & 90.8 (0.69) & 89.3 (0.73) & -- \\
Recall-Far & 30 & 66.7 (0.50) & 90.0 (0.78) & 66 & 90.9 (0.36) & 83.3 (0.59) & -- \\
Resist-Stale & 29 & 79.3 (0.68) & 82.8 (0.63) & 54 & 94.4 (0.70) & 82.8 (0.55) & -- \\
Resist-Intercut & 28 & 67.9 (0.49) & 89.3 (0.77) & 69 & 92.8 (0.67) & 89.3 (0.75) & -- \\
Resist-Lookalike & 30 & 73.3 (0.60) & 90.0 (0.78) & 80 & 85.0 (0.37) & 90.0 (0.77) & -- \\
Reuse-Own & 34 & 73.5 (0.60) & 91.2 (0.82) & 80 & 90.0 (0.38) & 91.2 (0.82) & -- \\
Reuse-Replaced & 33 & 54.5 (0.32) & 75.8 (0.45) & 48 & 89.6 (0.49) & 75.8 (0.45) & -- \\
\midrule
Gemini-3.1-Pro + Qwen-Edit$^\dagger$ & 48 & 75.0 (0.62) & 91.7 (0.83) & 119 & 92.4 (0.62) & 87.5 (0.74) & 41.7 / 37.5 \\
Nano Banana 2 \cite{nanobanana}$^\dagger$ & 50 & 70.0 (0.55) & 86.0 (0.72) & 112 & 92.9 (0.63) & 84.0 (0.66) & 42.0 / 34.0 \\
FLUX.3-klein \cite{flux3klein} & 48 & 70.8 (0.55) & 93.8 (0.86) & 106 & 89.6 (0.47) & 91.7 (0.81) & 31.2 / 31.2 \\
SenseNova-U1.5 \cite{sensenova} & 49 & 61.2 (0.43) & 83.7 (0.63) & 99 & 89.9 (0.49) & 81.6 (0.55) & 22.4 / 32.7 \\
BAGEL \cite{bagel} & 45 & 55.6 (0.32) & 75.6 (0.39) & 94 & 87.2 (0.58) & 86.7 (0.50) & 11.1 / 20.0 \\
\bottomrule
\end{tabular}\end{adjustbox}
\end{table*}

\paragraph{Follow.} The judge receives the two reference crops of $X$, the generated frame $\hat I_q$ and the name of
the prop. The crops come from the evidence frame of each twin, or for Reuse from the model's own frames. The judge
answers whether the prop in $\hat I_q$ matches crop 1, crop 2 or neither. The question is asked twice with the crops
in swapped order, and the side passes only if both answers name its own look. The judge samples at a temperature
of 0.2. In the prompt below,
\texttt{\{kind\}} stands for the name of the prop.
{\footnotesize
\begin{verbatim}
Images 1 and 2 are close-ups of two different
versions of {kind}. Image 3 is a scene. Which
version appears in image 3? Look at colours,
pattern, printed details and shape. Reply with
JSON only: {"answer": "1", "2" or "neither"}
\end{verbatim}}

\paragraph{Exec.} Each kept execution question is asked on $\hat I_q$ alone with yes or no answers, and a side passes
if all answers are correct. Questions never mention $X$, so Exec checks only the scripted change, and an unsure or missing answer counts as wrong. The prompt is:
{\footnotesize
\begin{verbatim}
Answer each question about this image with
"yes", "no" or "unsure" (unsure only if it truly
cannot be seen). Questions:
{numbered questions}
Return {"answers": ["yes"|"no"|"unsure", ...]}
with one answer per question, in order.
\end{verbatim}}

\paragraph{Human agreement.} We sample 250 side decisions with their 554 execution questions across subtypes and five systems. A human annotator sees the same crops and frames as the judge but not the system names or the
judge's answers, and labels 240 of the decisions and 530 of the questions. The annotator marks the other 10 decisions as unsure, and we leave them out. Within each system and subtype the sample mixes sides that the judge passes, sides with the twin's look and sides with neither look, so its pass rates differ from those of the full runs. Agreement on the side outcome is 86.2\%, with an interval of 81.9 to 90.6 and $\kappa = 0.68$. It is 86.2\% on Follow with $\kappa = 0.70$, 66.7\% on the three-way crop choice with $\kappa = 0.50$, and 90.6\% on execution questions with $\kappa = 0.56$. Most disagreements on the crop choice are between neither and one of the crops. A second annotator also labels 77 of the 240 decisions. The two annotators agree on 87.0\% of side outcomes with $\kappa = 0.62$, on 93.5\% of Follow with $\kappa = 0.83$ and on 89.6\% of crop choices with $\kappa = 0.79$. On the same decisions the judge agrees with the first annotator on 90.9\% of side outcomes and with the second on 85.7\%, and on the 67 decisions where the two annotators agree it matches them on 94.0\%. \cref{tab:human} breaks the agreement down by subtype and
system. Agreement is lowest on Reuse-Replaced, where the prop is often small or partly occluded, and the judge narrows the spread of pass rates between the five systems.

\section{Agents}
\label{app:agents}
\begin{table*}[tp]
\centering
\scriptsize
\setlength{\tabcolsep}{3pt}
\captionsetup{position=top}
\caption{\textbf{Planner outputs passed to the editor} on the 600 pairs of phase~A. All columns except Overall cover the sides with a stored plan. Words is the median length of the edit instruction, and $e$ shown is the share of sides on which $e$ is among the images given to the editor. Follow is split by whether $e$ was shown, with the number of sides without $e$ in the last column.}
\label{tab:agent}
\begin{adjustbox}{max width=\linewidth}\begin{tabular}{@{}l c c c cccccccc cc@{}}
\toprule
 & & & & \multicolumn{8}{c}{$e$ shown per subtype (\%)} & \multicolumn{2}{c}{Follow (\%)} \\
\cmidrule(lr){5-12}\cmidrule(l){13-14}
Agent & Overall & Words & $e$ shown & Start & Mid & Far & Stale & Intercut & Lookalike & Own & Replaced & $e$ shown & $e$ not shown \\
\midrule
GPT-5.6 + GPT Image 2.5 \cite{gptimage25}$^\dagger$ & 71.0 & 98 & 97.8 & 100 & 100 & 98 & 99 & 94 & 97 & 97 & 98 & 88.5 & 3.8 (26) \\
Qwen3.8-Max + Qwen-Edit$^\dagger$ & 60.9 & 109 & 98.2 & 100 & 99 & 99 & 100 & 97 & 97 & 99 & 95 & 84.5 & 4.5 (22) \\
Gemini-3.1-Pro + Qwen-Edit$^\dagger$ & 60.2 & 86 & 97.8 & 100 & 97 & 99 & 99 & 97 & 97 & 97 & 97 & 83.0 & 3.8 (26) \\
Kimi-K3 + Qwen-Edit$^\dagger$ & 56.7 & 97 & 98.2 & 100 & 100 & 99 & 99 & 94 & 98 & 98 & 97 & 82.9 & 0.0 (22) \\
GPT-5.6 + Qwen-Edit$^\dagger$ & 55.7 & 98 & 97.8 & 100 & 100 & 98 & 99 & 94 & 97 & 97 & 98 & 79.0 & 0.0 (26) \\
Gemini-3.1-Pro + Qwen-Image-2.1$^\dagger$ & 55.2 & 86 & 97.8 & 100 & 97 & 99 & 99 & 97 & 97 & 97 & 97 & 85.2 & 7.7 (26) \\
Kimi-K3 (no images) + Qwen-Edit$^\dagger$ & 42.2 & 95 & 89.4 & 96 & 100 & 75 & 100 & 65 & 83 & 99 & 97 & 71.9 & 7.1 (127) \\
Qwen3.8 + GPT Image 2.5$^\dagger$ & 29.3 & 91 & 60.4 & 70 & 71 & 42 & 61 & 42 & 55 & 86 & 57 & 71.6 & 5.1 (475) \\
Qwen3.8 + FLUX.3-klein & 20.2 & 91 & 60.4 & 70 & 71 & 42 & 61 & 42 & 55 & 86 & 57 & 63.0 & 5.5 (475) \\
Qwen3.5-9B + FLUX.3-klein & 19.7 & 70 & 69.5 & 77 & 77 & 40 & 85 & 44 & 62 & 87 & 84 & 58.6 & 5.7 (366) \\
Qwen3.8 + Qwen-Edit & 19.1 & 91 & 60.4 & 70 & 71 & 42 & 61 & 42 & 55 & 86 & 57 & 57.1 & 5.5 (475) \\
Qwen3.5-9B \cite{qwen35} + Qwen-Edit & 19.0 & 70 & 64.1 & 59 & 85 & 27 & 83 & 33 & 54 & 89 & 83 & 57.1 & 4.2 (431) \\
Qwen3.5-9B + FLUX.3-klein, no upsampler & 18.9 & 70 & 69.5 & 81 & 83 & 37 & 85 & 41 & 60 & 81 & 89 & 61.4 & 3.6 (366) \\
Qwen3.8 + Qwen-Image-2.1 & 17.6 & 99 & 46.2 & 41 & 75 & 33 & 49 & 23 & 24 & 79 & 47 & 65.4 & 5.3 (645) \\
\bottomrule
\end{tabular}\end{adjustbox}
\end{table*}

\cref{tab:agent} shows what each planner hands to the editor. Closed planners given the images pass the evidence
frame on 94 to 100\% of sides in every subtype. On the few sides where they omit it, the look is reproduced on at most 7.7\% of sides, so their remaining failures arise in the instruction and the editor. The open Qwen3.8 planner
passes the evidence frame on 60.4\% of sides with Qwen-Image-Edit and on 46.2\% with Qwen-Image-2.1, and on Recall-Far and Resist-Intercut on only 23 to 42\%. About half of its gap in Follow to the closed planners comes from these misses, since its agents follow on 57 to 65\% of sides given $e$ and on 5 to 6\% without it, and the rest comes from the instruction. The Kimi-K3 planner without images, which must name frames by index from the
scripts alone, still selects $e$ on 89.4\% of sides.

\paragraph{Editor swaps and agents of similar size.} An editor swap keeps the stored plans and replaces only the
editor. With the GPT-5.6 plans, GPT Image 2.5 receives $e$ on the same 97.8\% of sides as Qwen-Image-Edit, and on these sides Follow rises from 79.0\% to 88.5\%. With the Qwen3.8 plans, $e$ reaches the editor on 725 of the 1,200 sides, or 60.4\%. On these sides Follow rises from 57.1\% to 71.6\%, while on the others it stays
at 5 to 6\%. Behind GPT Image 2.5 the GPT-5.6 planner adds 41.6 points, with an interval of 36.9 to 46.3. The
interaction of planner and editor, the extra gain of each part when the other is the stronger one, is 5.1 points,
with an interval of $-$0.3 to 10.4.
In the open agents of \cref{tab:agent} with Qwen3.5-9B \cite{qwen35} or FLUX.3-klein, the editor also receives
only the frames that the planner selects, at most three. Replacing Qwen3.8 with Qwen3.5-9B or Qwen-Image-Edit with FLUX.3-klein changes
Overall by at most 1.1 points, and none of these differences excludes zero. Qwen3.5-9B passes $e$ more
often than Qwen3.8, on 64.1 to 69.5\% of sides against 60.4\%, but writes shorter instructions, with a median
of about 70 words against about 90. Given the full history without a planner, FLUX.3-klein scores 29.0\%, 9.4 points more than behind the Qwen3.5-9B planner. SenseNova-U1.5 has 17.5B parameters in total, fewer than the 19.7B of the smallest of these agents, Qwen3.5-9B with FLUX.3-klein and no upsampler, and with each system given the history in its own input format it scores 11.0 points more, with an interval of 6.6 to 15.6. After fine-tuning on \train it scores 30.6 points more. In these swap runs a locally served Qwen3.8-27B grounds the Reuse reference crops.

\section{\train details}
\label{app:train}
The chain counts of \cref{sec:train} include 5\% of each track held out from training.
\paragraph{Tracks.} Real revisits use keyframe chains from video groups not used by \bench. A step is scored only if
it returns to an earlier place, $X$-free same-place frames lie in between, and its DINOv2 similarity to the earlier
frame exceeds that to the previous frame. Intercut returns splice two synthetic chains, chain A, 1 to 6 shots of chain B and a return to A. The return script opens with one of six phrases that name the place of A again, such as \emph{Back at the}, \emph{Returning to the} or \emph{The story returns to the}. A synthetic script plans five steps from one of
80 places and one of 20 themes, with changes in time, space, cause or a combination. Each step describes the
resulting scene in 25 to 60 words without naming the colour, pattern or material of an object already visible, and
comes with yes or no questions about the change and about the objects that must stay the same. Qwen-Image-2.1 draws
frame 0 and edits each later step from the previous frame. Qwen3.8 accepts a step when at least two change questions
turn from no to yes and every kept object is still the same one. If three draws of a step fail, the remaining
steps are planned again from the current frame, at most twice per chain. Chains with fewer than two kept steps are
dropped. Verified edits are single steps of this pipeline that pass the same check \crefp{fig:train}.

\paragraph{Disjointness.}\label{app:disjoint} Train and benchmark are split by source video group. A DINOv2
nearest-neighbour search of all 11,472 distinct images in the 4,514 benchmark sides against the 131,830
training images gives a median similarity of 0.62, a 99th percentile of 0.89, and 0.035\% of images above 0.95. Per
pair, the most similar image has a median similarity of 0.81. Only two pairs have an image above 0.95, and both come
from MLVU videos that are near-duplicates of training videos in another MLVU group. Removing them leaves the score of every model trained on \train and of its base model unchanged to one decimal on the remaining 2,255 pairs. \train holds no training line from a source group of \bench.

\section{Fine-tuning details}
\label{app:ft}
\paragraph{Data.} All four models train on the shuffled four-track mixture of \train. Qwen-Image-Edit trains on a 32K subset of these targets with the same share of each track.

\paragraph{Optimisation.} All runs use AdamW with a 100-step warmup on eight GPUs per node. After the warmup the learning rate of SenseNova-U1.5 follows a cosine down to 10\% of its peak, and the other runs keep it constant. SenseNova-U1.5 and BAGEL are evaluated with EMA weights, with decays of 0.998 and 0.999. SenseNova-U1.5 uses one node, packs each GPU's batch to at most 16,384 tokens and uses no weight decay. BAGEL uses two nodes with about 32K tokens per GPU. Neither accumulates gradients. The LoRA runs use a learning rate of $10^{-4}$, one sample per GPU and adapters on the attention and MLP layers.

\paragraph{SenseNova-U1.5 and BAGEL.} Trained on the same four tracks, SenseNova-U1.5 gains 20.2 points and BAGEL 10.8 \crefp{tab:sft}. The two differ in size, recipe and starting score, so the comparison does not isolate one cause. SenseNova-U1.5 reads and draws pixel patches in one model with no vision encoder or VAE, while BAGEL reads each image with SigLIP and a VAE and draws in VAE latents. Before fine-tuning SenseNova-U1.5 already passes 45.9\% of Recall-Start pairs, against 7.2\% for BAGEL.

\paragraph{Reworded transitions.}\label{app:para} Training intercut returns open with phrases such as \emph{Back at the} or \emph{Returning to the}, and \bench scripts open with \emph{Back at}, \emph{Back in} or \emph{Cut to}. We rewrite only this opening phrase of
every phase-A script into one of six neutral wordings such as \emph{The action moves to}, \emph{We are now in} or
\emph{The next shot shows}, with the same wording for both twins, and keep the rest of each script, the frames and
the questions. Overall falls by 5.9 points for SenseNova-U1.5, from 29.8 to 23.9, and by 5.7 after fine-tuning, from 49.4 to 43.7. The new wording lowers both by a similar amount, and the fine-tuned model stays about 20 points ahead.

\section{Follow and Exec}
\label{app:fe}
\begin{table}[tp]
\centering
\scriptsize
\setlength{\tabcolsep}{2pt}
\captionsetup{position=top}
\caption{\textbf{Failure composition} in \% of sides. A failing side follows its own look but misses the change, shows the twin's look in both crop orders, shows neither, or, in Reuse, loses the prop in the frame it draws at $e$, which leaves no crop to compare. Open systems are scored on the full \bench and closed systems on phase~A, and unscored sides are left out.}
\label{tab:mix}
\begin{adjustbox}{max width=\linewidth}\begin{tabular}{@{}l ccccccc@{}}
\toprule
 & & & \multicolumn{2}{c}{Twin's look} & \multicolumn{2}{c}{Neither look} & \\
\cmidrule(lr){4-5}\cmidrule(lr){6-7}
System & Pass & Fol., no Exec & Exec & no Exec & Exec & no Exec & Lost at $e$ \\
\midrule
\multicolumn{8}{@{}l}{\textit{Reference (full)}} \\
Copy frame 0 & 0.5 & 54.5 & 0.0 & 8.8 & 0.1 & 21.1 & 14.9 \\
Copy previous frame & 0.0 & 1.1 & 0.3 & 9.5 & 1.0 & 65.0 & 22.9 \\
\midrule
\multicolumn{8}{@{}l}{\textit{Unified multimodal models (full)}} \\
SenseNova-U1.5 \cite{sensenova} & 41.4 & 4.9 & 4.5 & 0.6 & 39.5 & 6.3 & 2.9 \\
Emu3.5 \cite{emu35} & 31.0 & 24.4 & 3.0 & 1.7 & 19.2 & 14.0 & 6.7 \\
BAGEL \cite{bagel} & 13.2 & 8.3 & 3.6 & 2.2 & 40.0 & 26.7 & 6.0 \\
LLaDA2-Uni \cite{llada2uni} & 6.5 & 4.7 & 1.8 & 1.7 & 35.4 & 38.7 & 10.8 \\
HiDream-O1 \cite{hidreamo1} & 4.8 & 5.7 & 1.9 & 2.5 & 28.6 & 45.3 & 11.2 \\
OmniGen2 \cite{omnigen2} & 2.7 & 10.8 & 1.0 & 3.9 & 10.4 & 53.7 & 17.3 \\
Lumina-DiMOO \cite{lumina} & 0.6 & 13.5 & 0.3 & 8.1 & 1.4 & 55.5 & 20.5 \\
Show-o2 \cite{showo2} & 0.2 & 1.2 & 0.2 & 1.4 & 8.8 & 65.7 & 22.4 \\
Janus-Pro \cite{januspro} & 0.0 & 1.0 & 0.0 & 0.7 & 1.0 & 70.1 & 26.9 \\
\midrule
\multicolumn{8}{@{}l}{\textit{Image generators and editors (full)}} \\
FLUX.3-klein \cite{flux3klein} & 46.2 & 8.9 & 5.3 & 1.0 & 28.7 & 5.1 & 4.7 \\
VINCIE \cite{vincie} & 8.9 & 10.0 & 2.4 & 3.1 & 24.6 & 39.2 & 11.7 \\
FLUX.2-klein \cite{flux2} & 8.3 & 8.0 & 3.0 & 3.1 & 35.2 & 36.6 & 5.9 \\
Qwen-Image-2.1 \cite{qwenimage} & 6.7 & 4.7 & 2.9 & 2.1 & 36.0 & 35.6 & 11.7 \\
FLUX.2-dev \cite{flux2} & 6.2 & 4.4 & 2.3 & 2.0 & 37.8 & 40.2 & 7.1 \\
FireRed-Edit \cite{firered} & 5.2 & 27.5 & 1.0 & 5.0 & 8.3 & 43.9 & 9.0 \\
Qwen-Image-Edit \cite{qwenimage} & 3.3 & 15.5 & 0.7 & 2.8 & 10.8 & 56.4 & 10.4 \\
FLUX.1-Kontext \cite{fluxkontext} & 0.6 & 1.9 & 0.7 & 2.7 & 17.3 & 61.9 & 14.8 \\
\midrule
\multicolumn{8}{@{}l}{\textit{Agents (open planner + editor) (full)}} \\
Qwen3.8 + Qwen-Edit & 32.5 & 5.8 & 6.3 & 1.4 & 40.7 & 8.9 & 4.3 \\
Qwen3.8 + Qwen-Image-2.1 & 28.8 & 6.4 & 5.1 & 1.3 & 41.7 & 11.7 & 5.1 \\
\midrule
\multicolumn{8}{@{}l}{\textit{Keyframe-conditioned video models (full)}} \\
LTX-2.5 \cite{ltxvideo} & 19.8 & 11.1 & 3.1 & 1.8 & 31.0 & 21.7 & 11.5 \\
Wan-VACE \cite{vace} & 4.1 & 7.5 & 1.0 & 3.7 & 14.9 & 52.3 & 16.3 \\
StoryMem \cite{storymem} & 4.1 & 3.3 & 2.3 & 2.4 & 35.5 & 42.7 & 9.6 \\
HunyuanVideo-I2V \cite{hunyuanvideo} & 2.0 & 2.8 & 2.8 & 4.8 & 25.7 & 50.9 & 11.0 \\
Wan-I2V \cite{wan} & 1.6 & 1.6 & 2.2 & 2.4 & 35.6 & 46.5 & 10.1 \\
Wan-TI2V \cite{wan} & 0.8 & 1.3 & 1.7 & 3.5 & 22.1 & 52.2 & 18.5 \\
\midrule
\multicolumn{8}{@{}l}{\textit{Fine-tuned on \train (full)}} \\
\rowcolor{ours}SenseNova-U1.5 + \train & 61.9 & 9.3 & 1.6 & 0.3 & 19.0 & 3.8 & 4.1 \\
\rowcolor{ours}BAGEL + \train & 29.9 & 11.9 & 3.0 & 1.2 & 33.4 & 15.6 & 5.0 \\
\rowcolor{ours}Qwen-Image-2.1 + \train (LoRA) & 25.9 & 24.2 & 1.6 & 1.9 & 13.8 & 22.9 & 9.8 \\
\rowcolor{ours}Qwen-Image-Edit + \train (LoRA) & 23.8 & 28.6 & 1.5 & 2.5 & 10.7 & 25.9 & 7.0 \\
\midrule
\multicolumn{8}{@{}l}{\textit{Closed systems and agents (phase~A)}} \\
GPT-5.6 + GPT Image 2.5 \cite{gptimage25}$^\dagger$ & 80.3 & 6.2 & 1.9 & 0.4 & 8.2 & 1.0 & 1.7 \\
Qwen3.8-Max + Qwen-Edit$^\dagger$ & 73.8 & 9.1 & 1.6 & 0.1 & 10.4 & 1.8 & 3.0 \\
Gemini-3.1-Pro + Qwen-Edit$^\dagger$ & 72.0 & 9.2 & 1.6 & 0.2 & 12.9 & 1.9 & 2.0 \\
Kimi-K3 + Qwen-Edit$^\dagger$ & 70.7 & 10.6 & 2.2 & 0.3 & 12.8 & 1.9 & 1.3 \\
GPT-5.6 + Qwen-Edit$^\dagger$ & 69.7 & 7.5 & 2.5 & 0.4 & 15.2 & 1.8 & 2.8 \\
Gemini-3.1-Pro + Qwen-Image-2.1$^\dagger$ & 69.3 & 14.1 & 1.2 & 0.5 & 10.0 & 1.0 & 3.5 \\
Nano Banana 2 \cite{nanobanana}$^\dagger$ & 66.8 & 8.7 & 4.1 & 0.4 & 16.3 & 2.2 & 1.5 \\
Seedream 5 \cite{seedream5}$^\dagger$ & 60.5 & 7.7 & 6.8 & 1.0 & 17.0 & 3.3 & 3.4 \\
Nano Banana Pro \cite{nanobananapro}$^\dagger$ & 58.1 & 7.8 & 6.2 & 0.4 & 23.2 & 2.2 & 2.2 \\
Kimi-K3 (no images) + Qwen-Edit$^\dagger$ & 57.5 & 7.5 & 3.3 & 0.7 & 23.9 & 4.6 & 2.5 \\
Qwen3.8 + GPT Image 2.5$^\dagger$ & 40.9 & 4.3 & 9.7 & 1.0 & 36.0 & 4.2 & 3.8 \\
\bottomrule
\end{tabular}\end{adjustbox}
\end{table}

\cref{tab:mix} splits the failing sides further. Strong open models fail mostly by executing the change while
showing neither look, SenseNova-U1.5 on 39.5\% of sides. Every system shows the other twin's look on at most
10.7\% of sides, so most failures lose the look rather than swap it for the twin's. In Reuse a side
also fails when the frame the model drew at $e$ no longer shows the prop, on 1.3 to 3.8\% of sides for the closed
systems and up to 26.9\% for Janus-Pro. Copy frame 0 follows on more than half of its sides but almost never executes the change.

\paragraph{Look preference.}\label{app:look} Each look is the answer on one side of a pair and the wrong look on
the other, so a generator or a judge that prefers one look passes at most one side. Scored on the A sides alone and on the B sides alone, the 31 runs on the full \bench in \cref{tab:mainfull}, the four fine-tuned models and the two copy baselines included, rank with Kendall $\tau=0.98$ and differ by at most 1.6 points, so neither set of looks is favoured on average. Within a pair the two sides differ more often. The three runs above 20\% Overall pass exactly one side of 23.9 to 27.8\% of pairs, and on 2.9 to 6.8\% of pairs both of their outputs show the same look.
A benchmark with one history per item would count one of these two outputs as correct, although its look did not
come from its history.

\section{Evidence distance}
\label{sec:app-dist}
\begin{table}[tp]
\centering
\scriptsize
\setlength{\tabcolsep}{2.2pt}
\captionsetup{position=top}
\caption{\textbf{Pass rate by evidence distance} on the full \bench, binned by the number of shots between $e$ and $q$. Bins with fewer than 15 pairs are omitted.}
\label{tab:dist}
\begin{adjustbox}{max width=\linewidth}\begin{tabular}{@{}l ccc cccc cc@{}}
\toprule
Shots $q-e$ & \multicolumn{3}{c}{\textcolor{recall}{Recall}} & \multicolumn{4}{c}{\textcolor{resist}{Resist}} & \multicolumn{2}{c}{\textcolor{reuse}{Reuse}} \\
\cmidrule(lr){2-4}\cmidrule(lr){5-8}\cmidrule(lr){9-10}
 & 2 & 3--4 & 8 & 2 & 3--4 & 5--7 & 8 & 2 & 3--4 \\
Pairs & 394 & 454 & 221 & 187 & 232 & 133 & 27 & 548 & 61 \\
\midrule
\rowcolor{ours}SenseNova-U1.5 + \train & 46.4 & 52.0 & 56.1 & 51.3 & 54.7 & 51.9 & 37.0 & 44.9 & 34.4 \\
FLUX.3-klein \cite{flux3klein} & 34.0 & 46.0 & 37.6 & 8.6 & 36.2 & 43.6 & 51.9 & 19.7 & 36.1 \\
SenseNova-U1.5 \cite{sensenova} & 39.3 & 44.3 & 41.2 & 2.1 & 40.1 & 42.9 & 33.3 & 6.9 & 26.2 \\
Qwen3.8 + Qwen-Edit & 24.4 & 23.3 & 11.8 & 18.2 & 15.1 & 9.8 & 0.0 & 20.3 & 34.4 \\
Qwen3.8 + Qwen-Image-2.1 & 21.3 & 14.8 & 7.2 & 17.6 & 6.9 & 7.5 & 11.1 & 18.1 & 29.5 \\
Emu3.5 \cite{emu35} & 21.8 & 24.7 & 14.9 & 9.6 & 15.9 & 24.1 & 14.8 & 12.0 & 14.8 \\
\rowcolor{ours}BAGEL + \train & 18.8 & 23.6 & 14.5 & 11.2 & 18.5 & 15.8 & 11.1 & 14.2 & 11.5 \\
\rowcolor{ours}Qwen-Image-2.1 + \train (LoRA) & 17.8 & 16.1 & 6.3 & 15.0 & 13.4 & 15.0 & 0.0 & 11.9 & 0.0 \\
\rowcolor{ours}Qwen-Image-Edit + \train (LoRA) & 26.1 & 15.4 & 5.9 & 8.6 & 17.2 & 9.8 & 7.4 & 9.5 & 0.0 \\
LTX-2.5 \cite{ltxvideo} & 10.9 & 18.3 & 7.7 & 5.9 & 11.2 & 11.3 & 0.0 & 3.6 & 6.6 \\
BAGEL \cite{bagel} & 5.6 & 7.7 & 2.3 & 2.1 & 3.4 & 3.8 & 0.0 & 5.3 & 1.6 \\
FLUX.2-klein \cite{flux2} & 5.6 & 2.9 & 0.0 & 1.1 & 3.4 & 0.0 & 0.0 & 5.1 & 1.6 \\
VINCIE \cite{vincie} & 3.8 & 6.8 & 0.5 & 3.7 & 3.9 & 5.3 & 0.0 & 2.6 & 0.0 \\
Qwen-Image-2.1 \cite{qwenimage} & 3.6 & 2.6 & 0.0 & 0.5 & 0.9 & 5.3 & 3.7 & 2.4 & 1.6 \\
FLUX.2-dev \cite{flux2} & 3.3 & 1.8 & 0.0 & 1.1 & 1.7 & 1.5 & 0.0 & 1.8 & 4.9 \\
HiDream-O1 \cite{hidreamo1} & 3.3 & 2.9 & 0.0 & 0.0 & 1.7 & 0.8 & 0.0 & 1.5 & 1.6 \\
FireRed-Edit \cite{firered} & 6.6 & 0.9 & 0.5 & 0.5 & 0.4 & 0.0 & 0.0 & 3.1 & 0.0 \\
LLaDA2-Uni \cite{llada2uni} & 4.3 & 1.5 & 0.0 & 2.1 & 0.4 & 0.0 & 0.0 & 2.0 & 0.0 \\
Qwen-Image-Edit \cite{qwenimage} & 3.3 & 0.7 & 0.0 & 0.5 & 0.0 & 0.0 & 0.0 & 0.9 & 0.0 \\
Wan-VACE \cite{vace} & 3.3 & 1.1 & 0.0 & 1.6 & 0.0 & 0.0 & 0.0 & 0.5 & 0.0 \\
OmniGen2 \cite{omnigen2} & 0.3 & 0.7 & 0.0 & 3.2 & 0.4 & 0.0 & 0.0 & 0.5 & 0.0 \\
StoryMem \cite{storymem} & 0.8 & 0.2 & 0.0 & 1.6 & 0.0 & 0.0 & 0.0 & 0.7 & 0.0 \\
Lumina-DiMOO \cite{lumina} & 0.0 & 0.7 & 0.0 & 0.0 & 0.0 & 0.8 & 0.0 & 0.0 & 0.0 \\
HunyuanVideo-I2V \cite{hunyuanvideo} & 0.3 & 0.2 & 0.0 & 0.0 & 0.0 & 0.0 & 0.0 & 0.4 & 0.0 \\
Janus-Pro \cite{januspro} & 0.0 & 0.0 & 0.0 & 0.0 & 0.0 & 0.0 & 0.0 & 0.0 & 0.0 \\
Show-o2 \cite{showo2} & 0.0 & 0.0 & 0.0 & 0.0 & 0.0 & 0.0 & 0.0 & 0.0 & 0.0 \\
FLUX.1-Kontext \cite{fluxkontext} & 0.0 & 0.0 & 0.0 & 0.0 & 0.0 & 0.0 & 0.0 & 0.0 & 0.0 \\
Wan-TI2V \cite{wan} & 0.0 & 0.2 & 0.0 & 0.0 & 0.0 & 0.0 & 0.0 & 0.0 & 0.0 \\
Wan-I2V \cite{wan} & 0.0 & 0.0 & 0.0 & 0.0 & 0.0 & 0.0 & 0.0 & 0.2 & 0.0 \\
\bottomrule
\end{tabular}\end{adjustbox}
\end{table}

\cref{tab:dist} bins the full benchmark by the number of shots between the evidence frame $e$ and the query $q$.
SenseNova-U1.5 passes 39.3\%, 44.3\% and 41.2\% of Recall pairs at gaps of 2, 3 to 4 and 8 shots. Resist pairs at a
gap of two shots are all Resist-Stale, which explains their low rate of 2.1\%, and at larger gaps it passes
33 to 43\%. Within this range the subtype matters more than the distance.

\section{All systems on phase~A}
\label{app:phasea}
\begin{table*}[tp]
\centering
\scriptsize
\setlength{\tabcolsep}{3.2pt}
\captionsetup{position=top}
\caption{\textbf{All systems on phase~A}, 600 pairs with 75 per subtype and the columns of \cref{tab:main}. $^\dagger$ marks a closed model or planner. The controls give Qwen-Image-Edit only the previous frame with $s_q$ as the instruction, or $e$ and the previous frame with the full story, or render a prompt written from the scripts alone \crefp{sec:main,app:build}.}
\label{tab:phasea}
\begin{adjustbox}{max width=\linewidth}\begin{tabular}{@{}l c ccc ccc ccc cc cc@{}}
\toprule
 & & \multicolumn{3}{c}{Groups} & \multicolumn{3}{c}{\textcolor{recall}{Recall}} & \multicolumn{3}{c}{\textcolor{resist}{Resist}} & \multicolumn{2}{c}{\textcolor{reuse}{Reuse}} & \multicolumn{2}{c}{Diagnostic} \\
\cmidrule(lr){3-5}\cmidrule(lr){6-8}\cmidrule(lr){9-11}\cmidrule(lr){12-13}\cmidrule(l){14-15}
System & Overall & Recall & Resist & Reuse & Start & Mid & Far & Stale & Intercut & Lookalike & Own & Replaced & Follow & Exec \\
\midrule
\multicolumn{15}{@{}l}{\textit{Reference}} \\
Ground-truth frames & 100.0 & 100.0 & 100.0 & 100.0 & 100.0 & 100.0 & 100.0 & 100.0 & 100.0 & 100.0 & 100.0 & 100.0 & 100.0 & 100.0 \\
Copy frame 0 & 0.1 & 0.0 & 0.4 & 0.0 & 0.0 & 0.0 & 0.0 & 0.0 & 1.3 & 0.0 & 0.0 & 0.0 & 50.3 & 1.2 \\
Copy previous frame & 0.0 & 0.0 & 0.0 & 0.0 & 0.0 & 0.0 & 0.0 & 0.0 & 0.0 & 0.0 & 0.0 & 0.0 & 1.6 & 1.8 \\
\midrule
\multicolumn{15}{@{}l}{\textit{Unified multimodal models}} \\
SenseNova-U1.5 \cite{sensenova} & 29.8 & 40.4 & 31.1 & 18.0 & 45.3 & 33.3 & 42.7 & 4.0 & 52.0 & 37.3 & 28.0 & 8.0 & 47.6 & 88.5 \\
Emu3.5 \cite{emu35} & 15.6 & 17.8 & 15.6 & 13.3 & 16.0 & 18.7 & 18.7 & 10.7 & 21.3 & 14.7 & 16.0 & 10.7 & 55.4 & 56.0 \\
BAGEL \cite{bagel} & 5.0 & 3.5 & 3.6 & 8.0 & 5.3 & 5.3 & 0.0 & 4.0 & 4.0 & 2.7 & 4.0 & 12.0 & 18.4 & 60.2 \\
HiDream-O1 \cite{hidreamo1} & 1.1 & 0.9 & 0.4 & 2.0 & 1.3 & 1.3 & 0.0 & 0.0 & 1.3 & 0.0 & 2.7 & 1.3 & 11.2 & 37.4 \\
LLaDA2-Uni \cite{llada2uni} & 0.9 & 0.4 & 0.9 & 1.4 & 1.3 & 0.0 & 0.0 & 1.3 & 0.0 & 1.3 & 0.0 & 2.7 & 8.7 & 45.0 \\
OmniGen2 \cite{omnigen2} & 0.6 & 0.4 & 1.3 & 0.0 & 1.3 & 0.0 & 0.0 & 2.7 & 0.0 & 1.3 & 0.0 & 0.0 & 16.6 & 16.2 \\
Lumina-DiMOO \cite{lumina} & 0.1 & 0.0 & 0.4 & 0.0 & 0.0 & 0.0 & 0.0 & 0.0 & 1.3 & 0.0 & 0.0 & 0.0 & 10.2 & 3.3 \\
Janus-Pro \cite{januspro} & 0.0 & 0.0 & 0.0 & 0.0 & 0.0 & 0.0 & 0.0 & 0.0 & 0.0 & 0.0 & 0.0 & 0.0 & 1.2 & 0.8 \\
Show-o2 \cite{showo2} & 0.0 & 0.0 & 0.0 & 0.0 & 0.0 & 0.0 & 0.0 & 0.0 & 0.0 & 0.0 & 0.0 & 0.0 & 1.8 & 10.7 \\
\midrule
\multicolumn{15}{@{}l}{\textit{Image generators and editors}} \\
FLUX.3-klein \cite{flux3klein} & 29.0 & 36.0 & 27.1 & 24.0 & 34.7 & 26.7 & 46.7 & 8.0 & 34.7 & 38.7 & 26.7 & 21.3 & 54.1 & 84.5 \\
VINCIE \cite{vincie} & 4.1 & 4.0 & 4.4 & 4.0 & 12.0 & 0.0 & 0.0 & 5.3 & 2.7 & 5.3 & 0.0 & 8.0 & 19.1 & 37.1 \\
FLUX.2-klein \cite{flux2} & 3.0 & 2.2 & 2.2 & 4.7 & 5.3 & 1.3 & 0.0 & 0.0 & 1.3 & 5.3 & 4.0 & 5.3 & 15.9 & 49.1 \\
Qwen-Image-2.1 \cite{qwenimage} & 2.5 & 0.9 & 2.7 & 4.0 & 2.7 & 0.0 & 0.0 & 0.0 & 5.3 & 2.7 & 5.3 & 2.7 & 12.1 & 51.6 \\
FLUX.2-dev \cite{flux2} & 2.2 & 2.7 & 1.8 & 2.0 & 5.3 & 2.7 & 0.0 & 1.3 & 2.7 & 1.3 & 4.0 & 0.0 & 11.1 & 48.0 \\
FireRed-Edit \cite{firered} & 1.9 & 1.3 & 0.4 & 4.0 & 2.7 & 0.0 & 1.3 & 1.3 & 0.0 & 0.0 & 1.3 & 6.7 & 34.8 & 11.2 \\
Qwen-Image-Edit \cite{qwenimage} & 1.0 & 0.4 & 0.0 & 2.6 & 1.3 & 0.0 & 0.0 & 0.0 & 0.0 & 0.0 & 4.0 & 1.3 & 21.3 & 11.5 \\
FLUX.1-Kontext \cite{fluxkontext} & 0.0 & 0.0 & 0.0 & 0.0 & 0.0 & 0.0 & 0.0 & 0.0 & 0.0 & 0.0 & 0.0 & 0.0 & 3.1 & 21.1 \\
\midrule
\multicolumn{15}{@{}l}{\textit{Agents (open planner + editor)}} \\
Qwen3.8 + Qwen-Edit & 19.1 & 20.4 & 14.2 & 22.6 & 24.0 & 21.3 & 16.0 & 17.3 & 10.7 & 14.7 & 28.0 & 17.3 & 36.7 & 83.3 \\
Qwen3.8 + Qwen-Image-2.1 & 17.6 & 19.6 & 9.8 & 23.3 & 14.7 & 34.7 & 9.3 & 18.7 & 4.0 & 6.7 & 33.3 & 13.3 & 33.1 & 80.5 \\
\midrule
\multicolumn{15}{@{}l}{\textit{Keyframe-conditioned video models}} \\
LTX-2.5 \cite{ltxvideo} & 7.5 & 9.3 & 9.8 & 3.4 & 10.7 & 9.3 & 8.0 & 2.7 & 14.7 & 12.0 & 4.0 & 2.7 & 26.0 & 61.4 \\
Wan-VACE \cite{vace} & 0.9 & 1.3 & 1.3 & 0.0 & 1.3 & 2.7 & 0.0 & 4.0 & 0.0 & 0.0 & 0.0 & 0.0 & 10.1 & 20.9 \\
StoryMem \cite{storymem} & 0.8 & 0.0 & 0.9 & 1.4 & 0.0 & 0.0 & 0.0 & 2.7 & 0.0 & 0.0 & 0.0 & 2.7 & 6.9 & 42.8 \\
HunyuanVideo-I2V \cite{hunyuanvideo} & 0.2 & 0.0 & 0.0 & 0.7 & 0.0 & 0.0 & 0.0 & 0.0 & 0.0 & 0.0 & 1.3 & 0.0 & 5.4 & 33.3 \\
Wan-TI2V \cite{wan} & 0.1 & 0.4 & 0.0 & 0.0 & 0.0 & 1.3 & 0.0 & 0.0 & 0.0 & 0.0 & 0.0 & 0.0 & 2.7 & 26.1 \\
Wan-I2V \cite{wan} & 0.0 & 0.0 & 0.0 & 0.0 & 0.0 & 0.0 & 0.0 & 0.0 & 0.0 & 0.0 & 0.0 & 0.0 & 3.7 & 40.7 \\
\midrule
\multicolumn{15}{@{}l}{\textit{Closed image generators}} \\
Nano Banana 2 \cite{nanobanana}$^\dagger$ & 50.4 & 57.8 & 53.3 & 40.0 & 62.7 & 48.0 & 62.7 & 29.3 & 68.0 & 62.7 & 33.3 & 46.7 & 75.4 & 88.5 \\
Seedream 5 \cite{seedream5}$^\dagger$ & 45.7 & 57.8 & 46.7 & 32.6 & 60.0 & 50.7 & 62.7 & 37.3 & 48.0 & 54.7 & 32.0 & 33.3 & 68.2 & 87.5 \\
Nano Banana Pro \cite{nanobananapro}$^\dagger$ & 39.3 & 42.2 & 40.4 & 35.4 & 42.7 & 32.0 & 52.0 & 29.3 & 48.0 & 44.0 & 32.0 & 38.7 & 65.8 & 89.4 \\
\midrule
\multicolumn{15}{@{}l}{\textit{Agents with a closed planner or editor}} \\
GPT-5.6 + GPT Image 2.5 \cite{gptimage25}$^\dagger$ & 71.0 & 75.1 & 73.3 & 64.7 & 76.0 & 78.7 & 70.7 & 78.7 & 72.0 & 69.3 & 52.0 & 77.3 & 86.7 & 92.3 \\
Qwen3.8-Max + Qwen-Edit$^\dagger$ & 60.9 & 64.0 & 64.0 & 54.7 & 70.7 & 57.3 & 64.0 & 66.7 & 61.3 & 64.0 & 46.7 & 62.7 & 83.0 & 88.7 \\
Gemini-3.1-Pro + Qwen-Edit$^\dagger$ & 60.2 & 62.7 & 65.3 & 52.6 & 68.0 & 56.0 & 64.0 & 70.7 & 61.3 & 64.0 & 44.0 & 61.3 & 81.2 & 88.4 \\
Kimi-K3 + Qwen-Edit$^\dagger$ & 56.7 & 59.1 & 57.8 & 53.3 & 64.0 & 48.0 & 65.3 & 60.0 & 56.0 & 57.3 & 49.3 & 57.3 & 81.2 & 87.0 \\
GPT-5.6 + Qwen-Edit$^\dagger$ & 55.7 & 65.8 & 54.7 & 46.6 & 62.7 & 64.0 & 70.7 & 48.0 & 57.3 & 58.7 & 40.0 & 53.3 & 77.2 & 90.0 \\
Gemini-3.1-Pro + Qwen-Image-2.1$^\dagger$ & 55.2 & 61.4 & 62.2 & 42.0 & 62.7 & 54.7 & 66.7 & 62.7 & 60.0 & 64.0 & 37.3 & 46.7 & 83.5 & 83.8 \\
Kimi-K3 (no images) + Qwen-Edit$^\dagger$ & 42.2 & 46.7 & 41.3 & 38.6 & 56.0 & 46.7 & 37.3 & 42.7 & 32.0 & 49.3 & 33.3 & 44.0 & 65.0 & 86.9 \\
Qwen3.8 + GPT Image 2.5$^\dagger$ & 29.3 & 36.0 & 25.3 & 26.7 & 44.0 & 44.0 & 20.0 & 29.3 & 18.7 & 28.0 & 38.7 & 14.7 & 45.2 & 90.0 \\
\midrule
\multicolumn{15}{@{}l}{\textit{Agents with a smaller open planner or another open editor}} \\
Qwen3.8 + FLUX.3-klein & 20.2 & 21.8 & 11.6 & 27.3 & 25.3 & 24.0 & 16.0 & 14.7 & 8.0 & 12.0 & 33.3 & 21.3 & 40.2 & 79.7 \\
Qwen3.5-9B + FLUX.3-klein & 19.7 & 20.5 & 16.0 & 22.6 & 22.7 & 22.7 & 16.0 & 12.0 & 17.3 & 18.7 & 21.3 & 24.0 & 42.5 & 76.0 \\
Qwen3.5-9B \cite{qwen35} + Qwen-Edit & 19.0 & 19.1 & 16.4 & 21.4 & 16.0 & 32.0 & 9.3 & 21.3 & 12.0 & 16.0 & 24.0 & 18.7 & 38.1 & 77.2 \\
Qwen3.5-9B + FLUX.3-klein, no upsampler & 18.9 & 20.0 & 13.3 & 23.4 & 29.3 & 20.0 & 10.7 & 10.7 & 12.0 & 17.3 & 22.7 & 24.0 & 43.8 & 70.6 \\
\midrule
\multicolumn{15}{@{}l}{\textit{Controls without a planner or without images}} \\
Text-only pipeline & 14.3 & 0.4 & 0.0 & 42.6 & 0.0 & 0.0 & 1.3 & 0.0 & 0.0 & 0.0 & 45.3 & 40.0 & -- & 87.7 \\
Qwen-Edit, $e$, previous frame and story & 2.7 & 1.8 & 0.9 & 5.3 & 1.3 & 4.0 & 0.0 & 1.3 & 1.3 & 0.0 & 6.7 & 4.0 & -- & 32.2 \\
Qwen-Edit, previous frame only & 0.4 & 0.0 & 0.4 & 0.7 & 0.0 & 0.0 & 0.0 & 1.3 & 0.0 & 0.0 & 1.3 & 0.0 & -- & 83.0 \\
\midrule
\multicolumn{15}{@{}l}{\textit{Fine-tuned on \train (ours)}} \\
\rowcolor{ours}SenseNova-U1.5 + \train & 49.4 & 46.2 & 57.3 & 44.7 & 56.0 & 25.3 & 57.3 & 62.7 & 57.3 & 52.0 & 34.7 & 54.7 & 70.1 & 87.3 \\
\rowcolor{ours}BAGEL + \train & 14.9 & 15.1 & 18.2 & 11.3 & 20.0 & 6.7 & 18.7 & 16.0 & 20.0 & 18.7 & 6.7 & 16.0 & 38.8 & 67.8 \\
\rowcolor{ours}Qwen-Image-2.1 + \train (LoRA) & 11.3 & 11.1 & 14.7 & 8.0 & 17.3 & 5.3 & 10.7 & 14.7 & 14.7 & 14.7 & 2.7 & 13.3 & 46.9 & 43.9 \\
\rowcolor{ours}Qwen-Image-Edit + \train (LoRA) & 8.7 & 10.2 & 11.1 & 4.7 & 20.0 & 6.7 & 4.0 & 6.7 & 20.0 & 6.7 & 1.3 & 8.0 & 49.8 & 33.1 \\
\bottomrule
\end{tabular}\end{adjustbox}
\end{table*}

\cref{tab:phasea} lists every system on the 600 phase-A pairs, the only split on which open and closed systems
share the same pairs. The 25 open systems rank on phase~A almost as on the full \bench, with Kendall $\tau = 0.95$,
and $\tau = 0.96$ when the four fine-tuned models are added.

\section{General generation after fine-tuning}
\label{app:general}
\begin{table}[tp]
\centering
\scriptsize
\setlength{\tabcolsep}{3pt}
\captionsetup{position=top}
\caption{\textbf{General generation and editing} of SenseNova-U1.5 before and after fine-tuning: GenEval \cite{geneval} with 553 prompts and GEdit-Bench \cite{step1xedit} with 606 edits, the latter judged by Qwen3.8. SC, PQ and O are semantic consistency, perceptual quality and overall.}
\label{tab:general}
\begin{adjustbox}{max width=\linewidth}\begin{tabular}{@{}l ccccccc ccc@{}}
\toprule
 & \multicolumn{7}{c}{GenEval (\% correct images)} & \multicolumn{3}{c}{GEdit-Bench} \\
\cmidrule(lr){2-8}\cmidrule(l){9-11}
Checkpoint & Single & Two & Count & Colors & Pos. & Attr. & All & SC & PQ & O \\
\midrule
SenseNova-U1.5 & 100.0 & 97.5 & 89.4 & 90.2 & 89.2 & 78.8 & 90.5 & 7.46 & 7.03 & 7.09 \\
+ \train & 100.0 & 97.5 & 86.6 & 93.3 & 88.5 & 80.8 & 90.9 & 7.50 & 6.08 & 6.57 \\
\bottomrule
\end{tabular}\end{adjustbox}
\end{table}

Fine-tuning on \train barely changes general text-to-image generation. On the 553 prompts of GenEval
\cite{geneval}, 90.5\% of the images of SenseNova-U1.5 are correct before fine-tuning and 90.9\% after. \cref{tab:general} gives GenEval per category, and each category changes by about 3 points at most, on colours and counting. Image editing from a single instruction changes more. The overall GEdit-Bench score falls from 7.09 to 6.57, mostly on perceptual quality.

\paragraph{Multi-turn editing data from elsewhere.} BAGEL fine-tuned on WEAVE-100k \cite{weave}, the multi-turn editing
data of WEAVE, passes 6.4\% of the phase-A pairs, against 5.0\% before, a difference of $+$1.4 with an interval of
$-$1.2 to 4.1. BAGEL fine-tuned on \train reaches 14.9\% \crefp{sec:sft}.

\section{Sources for the comparison table}
\label{app:tab1}
\begin{table*}[tp]
\centering
\scriptsize
\renewcommand{\arraystretch}{1.12}
\captionsetup{position=top}
\caption{\textbf{Sources for \cref{tab:compare}}, with the arXiv version we read and the sections behind each row. A
mark for a missing component means that a search of the full text found none.}
\label{tab:audit}
\begin{tabular}{@{}p{2.3cm}p{1.6cm}p{12.7cm}@{}}
\toprule
Benchmark & Version & Evidence for the marks in \cref{tab:compare} \\
\midrule
ViStoryBench & 2505.24862v5 & 80 multi-shot stories from scripts and literature, \S3.1--3.2. Multimodal models get the previous shot and character references, App.~F.5. Self-similarity over own outputs and VLM alignment, \S3.3. The copy-paste baseline of \S4.2 copies character references rather than a history frame. \\
DramaChain & 2609.00646v2 & Six-stage chain from 20 dramas, \S3.2. Inputs are reference sheets, not earlier shots, and each stage starts from a fixed upstream, \S1. Five-point rubric, \S3.3. Only judge-training data is built. \\
KeyFrame-Compass & 2607.14202v1 & Multi-shot samples given as an ordered multi-image list, \S3.1. One video per call. Keyframes are drawn by image generators, partly from captions of real videos, \S3.2. Keyframe matching by DINOv3 and PSNR/SSIM plus a VLM checklist, \S3.3. \\
LongMIG-Bench & 2610.17080v1 & 100 stories of about nine shots, \S4.1, App.~C.1. Context is the model's own earlier images plus at most one user photo, \S3.4. GPT judge on a 1--5 scale; a text-planned per-image pipeline scores high, Tab.~1. App.~B.2 builds a 21,162-story corpus but does not say whether it will be released. \\
VMem-Bench & 2610.35001v1 & Track-B rolls out 30 authored stories of five-second segments with prompts as input and references drawn from the model's own generated history, \S3.2, \S4.1. Track-A scores reference selection from real film frames before any generation, \S3.1. A no-memory control is scored and stays far from zero, Tab.~3, and no copy baseline is reported. VLM presence judge, \S3.3. No training set. \\
WEAVE & 2511.11434v1 & Multi-turn edits of one scene, \S3.1, with all previous interactions in context and self-generated history, \S4.1. VLM key-point scoring, \S3.2. Web-collected images, App.~A.2. WEAVE-100k used for training, \S4.2. \\
MSE-Bench & 2506.10941v2 & Five-turn editing sessions where turn $i$ edits the model's turn $i{-}1$ output, \S4.2--4.3. Images from MS-COCO and LAION, App.~C.6. No ground truth, GPT-4o scoring, \S4.2. VINCIE-10M released on the project page. \\
OCCUR-Bench & 2607.07051v1 & Two- to five-turn edits of one base image from COCO, PIE-Bench and HQ-Edit, \S3.1--3.2. Full image history for the orchestrator, Eq.~1--2. Restoration scored against annotated reference states plus an LLM judge, App.~A.4. Training-free. \\
IntReal & 2610.38425v1 & Real web and cartoon sequences in four categories, with 150 of the 600 test items being story episodes, \S2.1, \S3.1. The fine-grained setting conditions on real preceding text and images, and the holistic setting does not check later images against the model's own earlier ones, \S3.2, App.~B.2. Expert models and a VLM judge, with no shortcut baseline, \S4.1. IntReal-100K is used for SFT and RL and released, \S4.3, App.~A.1. \\
StEvo-Bench & 2603.13215v1 & One continuous video from an initial image and a prompt, \S3.1. State after occlusion judged against the model's own frames by a VLM, \S3.2. The source of the initial images is not stated. \\
WROP & 2609.28654v1 & 60-frame input clips rendered in Blender, \S3.2. Continuation from ground-truth input, \S4.2, scored by blind human preference, \S4.1. A 1.5M-sample training corpus is released. \\
MemoBench & 2606.27537v6 & Disappear-and-reappear clips from a first frame and a prompt, \S3.3. 164 of 360 clips are real, the rest rendered, \S3.2. Identity scored against the generated first frame, \S3.4, plus VLM questions, \S3.5. \\
R2M-Bench & 2608.27328v1 & One reference image and a navigation script, \S3. Revisits compared with samples from the same generated video, \S4.1. Gap-matched non-revisits calibrate the metric and are not a reported shortcut baseline, \S4.2. VLM rubric and human ratings, \S4.3. \\
VBVR-Pro & 2608.26105v2 & 300 procedural tasks with solvers and verifiable scorers, \S2.2, \S3.2. One question frame as input. Interleaved rollouts feed intermediate images back; removing the input image drops the score to 0.064, Tab.~6. All data released. \\
StateAliasBench & 2609.32679v1 & Pairs of real Android trajectories share the last screen and action but differ in hidden state, and each branch is scored as one next-screen prediction, \S2.1--2.2. The full screenshot history is an input and there is no rollout, App.~D.2. No copy or script-only baseline is scored, Tab.~2. A 2,538-example training set for the state estimator is described as released, App.~D.1. \\
WorldState-Bench & 2610.26811v1 & Fifty twin pairs, each twin one clip from a shared first frame plus rule text, \S4, App.~A. Multi-turn episodes appear only qualitatively, \S5.3. A VLM answers fixed-choice questions, \S4, App.~B, and the copy-paste rate counts identical verdicts across twins rather than scoring a frame-copy baseline, App.~D. No training set, \S5.1. \\
\bottomrule
\end{tabular}

\end{table*}

The columns of \cref{tab:compare} are defined as follows. \emph{Multi-shot}: items are sequences of shots of one
story, partly if they are multi-turn edits of one image. \emph{Image history}: several images are given as input,
partly if only one is. \emph{Own output}: later steps are scored on content the model drew earlier. \emph{Shortcuts fail}: the paper reports that copying a history frame and reading only the script both score near zero, partly if it reports only one of them or if they fail on only some of its tasks. \emph{Train set}: a training set is or will be released, partly if one is built but its release is not stated.
In the \bench row, Own output refers to the Reuse pairs. Copying frame 0 and reading only the script score near zero
on Recall and Resist, and Reuse leaves a textual route open by design \crefp{app:build}, so this cell is partial.
We filled \cref{tab:compare} from the latest arXiv version of each paper and checked every cell against its text on
2026-10-26, and the rows of IntReal, VMem-Bench, StateAliasBench and WorldState-Bench on 2026-11-08. \Cref{tab:audit} lists the version and the sections behind each row. Two rules decide the borderline
cells. Images that a benchmark gives as input count under Image history, and images the model drew itself count
under Own output, so LongMIG-Bench and VMem-Bench, whose context is mostly the model's own earlier images, are
partly under Image history. A shortcut counts as reported only if the paper scores it as a baseline, so a calibration term such as the
gap-matched non-revisits of R2M-Bench, or a qualitative remark about a model that copies the first frame, does not count.

\section{Data sources}
\label{app:sources}
The licences of Ego4D \cite{ego4d} and Ego-Exo4D \cite{egoexo4d} do not allow redistribution, so their keyframes
are not included in the release. Users obtain the two datasets from their official sites and run our script, which extracts these keyframes and rebuilds the pairs of
\bench and the chains of \train that use them. All other frames are included as images. The three smaller
egocentric sets of \cref{sec:construction} are EgoExoLearn \cite{egoexolearn}, TeleEgo \cite{teleego} and
SuperMemory-VQA \cite{supermemory}, and the real revisits of \train also use MovieChat-1K \cite{moviechat} and CASTLE
2024 \cite{castle}.

\section{Prompts}
\label{app:prompts}
All systems receive the same scripts and time gaps, and the same text is used for every pair, the
synthetic chains included.

\paragraph{Interleaved generators.} UMMs and multi-image generators receive this text, with frame 0 and the
earlier keyframes inserted as images at their \texttt{<image>} marks, within the image limits of \cref{app:models}.
{\footnotesize
\begin{verbatim}
You are drawing the next keyframe of a storyboard
made from real footage. The images given are
frame 0 and the earlier keyframes, in order. Each
keyframe follows a script that only states what
changed; anything the script does not change
(people, objects, places, their look) must stay
exactly as in the earlier frames.
Frame 0: <image>
Keyframe 1 (<time gap>): <script 1> <image>
...
Now draw ONLY Keyframe j (<time gap>): <script j>
Output one photorealistic frame with the same
camera style and aspect ratio as frame 0.
\end{verbatim}}
\paragraph{Multi-reference editors.} Editors get the shown frames as Picture 1, 2 and so on, and the story so far.
{\footnotesize
\begin{verbatim}
The input pictures are keyframes of a storyboard
made from real footage. Story so far:
Frame 0 (start) [Picture 1]
Keyframe 1 (<time gap>): <script 1> [not shown]
...
Now draw keyframe j (<time gap>): <script j>
The script only states what changed. Keep
everything it does not mention (people, objects,
place, lighting, camera style) as in Picture N,
and keep people, objects and places looking the
same as in the earlier keyframes. Output one
photorealistic frame.
\end{verbatim}}
\paragraph{Video models.} Keyframe-conditioned video models receive the scripts as one paragraph.
{\footnotesize
\begin{verbatim}
Earlier in this video: (<time gap>) <script 1>
... Now: <script j> The video ends on a frame
that shows exactly this state; everything the
description does not change stays the same.
\end{verbatim}}
\paragraph{Agent planner.} The planner receives the instruction below, then one line per history frame, each
followed by its image, and finally the request for the next keyframe. At most ten images are shown, frame 0 and the latest nine, which covers every history of \bench. A frame without its image keeps its line, which then ends with \texttt{(image not shown)}, and in the
variant without images this holds for every frame.
{\footnotesize
\begin{verbatim}
You direct an image-editing model that takes 1-3
input pictures and an instruction. You are shown
frame 0 and the earlier keyframes of a storyboard
made from real footage, each with the script that
produced it. Each script only states what
changed; anything it does not change must stay
exactly as in the earlier frames. The editor must
output the NEXT keyframe.
1. Choose the 1-3 images (by number) the editor
   should get: usually the frame that best shows
   the current scene, plus frames that show
   people/objects the new keyframe needs but that
   are hidden or changed later. The editor sees
   them as Picture 1, Picture 2, ... in the order
   you list them.
2. Write a self-contained edit instruction (at
   most 120 words, English) that refers to them
   as "Picture N": which picture is the base,
   what to change according to the new script,
   and what to keep or copy from which picture.
   The result must be one photorealistic frame of
   the same footage.
Reply as JSON:
{"frames": [numbers], "instruction": "..."}

Image 0: frame 0 (start).
<image>
Image 1: keyframe 1 (<time gap>): <script 1>
<image>
...
Next: keyframe j (<time gap>): <script j>
Choose the images (0-<j-1>) and write the
instruction for keyframe j.
\end{verbatim}}
\paragraph{Text-only pipeline.} Qwen3.8 receives this instruction and the scripts, and Qwen-Image-2512 draws an
image from the prompt that Qwen3.8 writes.
{\footnotesize
\begin{verbatim}
You write prompts for a text-to-image model.
Neither you nor the model can see any image. You
are given the scripts of a storyboard made from
real footage, one per keyframe, in order; each
script only states what changed. Whatever the
scripts say about a person, object or place (its
color, material, shape, pattern) stays true until
a later script changes it.
Write ONE self-contained, detailed prompt
(150-250 words, English) for the NEXT keyframe:
- apply the new script exactly;
- carry over everything the earlier scripts said
  that still holds, especially the stated look of
  every object;
- state the photographic style (real footage);
- never refer to "the image", "keyframe 2" or "as
  before": describe things directly.
Reply as JSON: {"prompt": "..."}

Frame 0: the opening frame (no script).
Keyframe 1 (<time gap>): <script 1>
...
Next: keyframe j (<time gap>): <script j>
Write the prompt for keyframe j.
\end{verbatim}}

\renewcommand{\dblfloatpagefraction}{0.9}
\section{Gallery}
\label{app:gallery}
\begin{figure*}[p]
\centering
\includegraphics[width=1.000\linewidth]{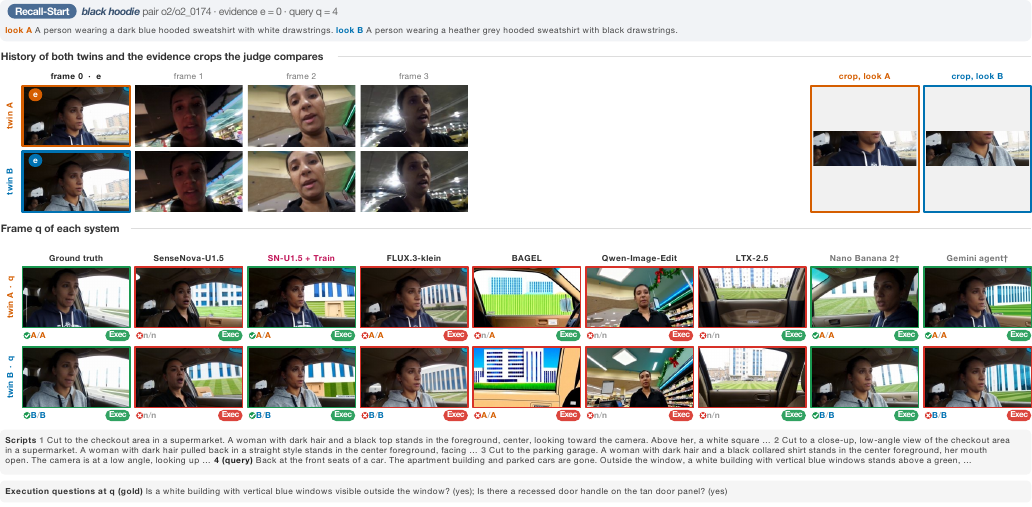}
\caption{\textbf{Recall-Start example.} Top: both twin histories, subsampled when long but keeping frame 0 and $e$, and the judge's two crops. Bottom: frame $q$ of each system for both twins, framed in green if the side passes Follow and Exec and in red if not. Below each output are the judge's answers in both crop orders, A, B or n for neither, and the Exec result. The pair was chosen automatically.}
\label{fig:viz-sub0}
\vspace{12pt}
\includegraphics[width=1.000\linewidth]{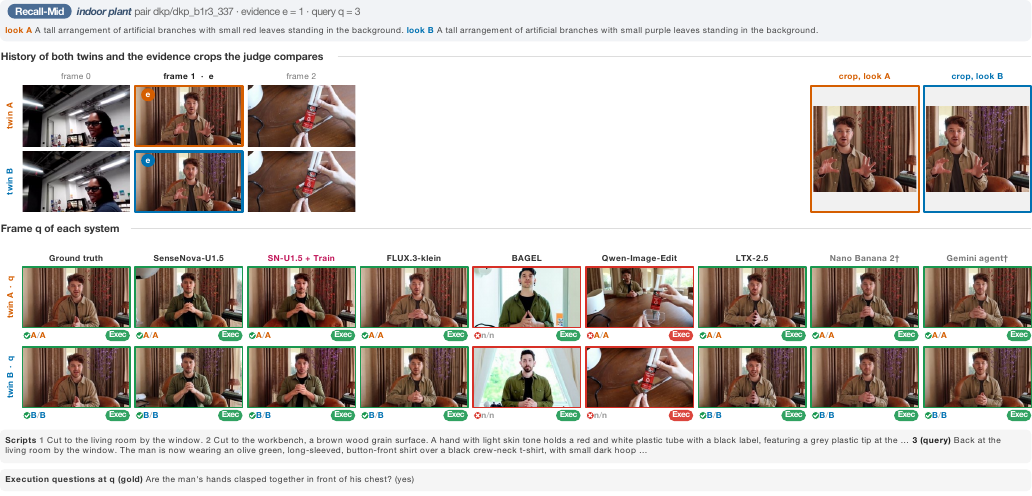}
\caption{\textbf{Recall-Mid example.} Layout and selection as in \cref{fig:viz-sub0}.}
\label{fig:viz-sub1}
\end{figure*}

\begin{figure*}[p]
\centering
\includegraphics[width=1.000\linewidth]{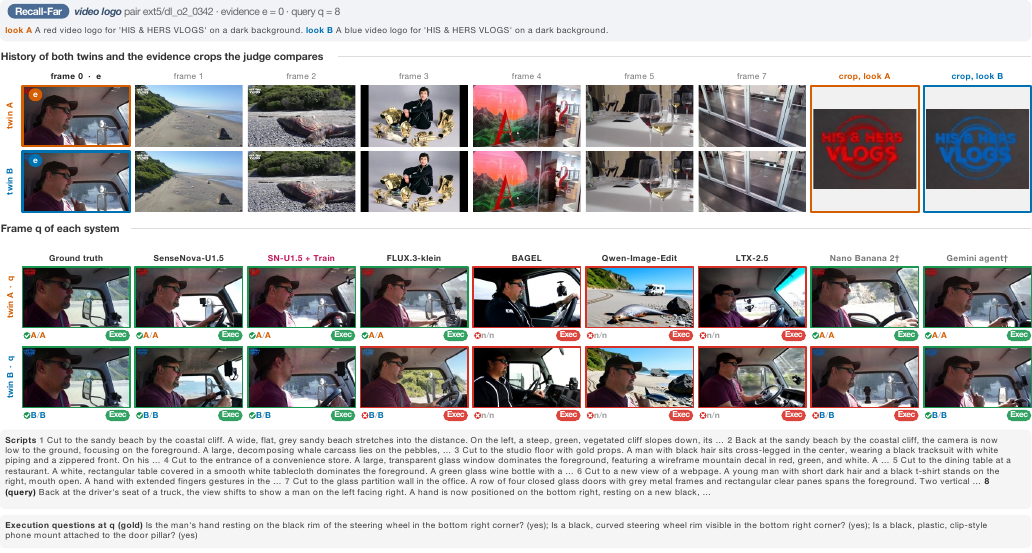}
\caption{\textbf{Recall-Far example.} Layout and selection as in \cref{fig:viz-sub0}.}
\label{fig:viz-sub2}
\vspace{12pt}
\includegraphics[width=1.000\linewidth]{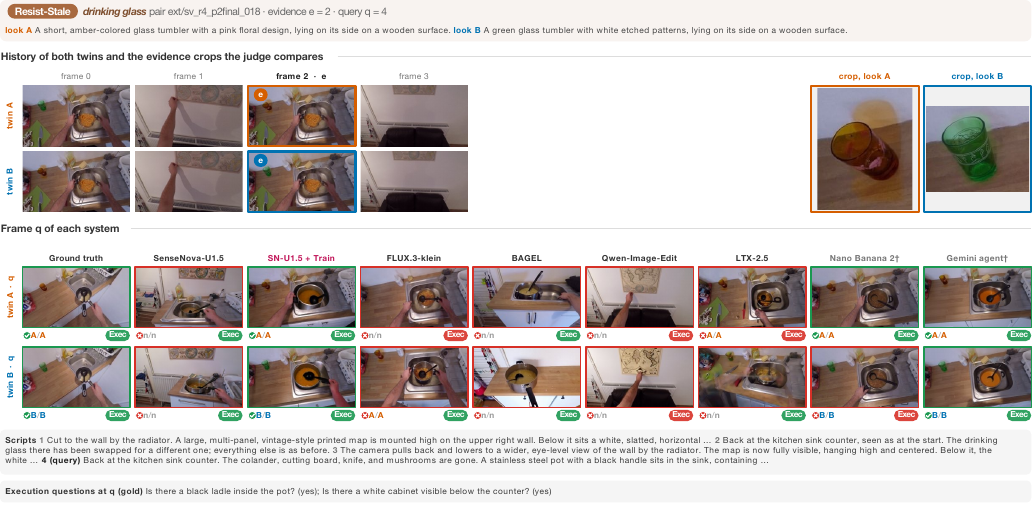}
\caption{\textbf{Resist-Stale example.} Layout and selection as in \cref{fig:viz-sub0}.}
\label{fig:viz-sub3}
\end{figure*}

\begin{figure*}[p]
\centering
\includegraphics[width=1.000\linewidth]{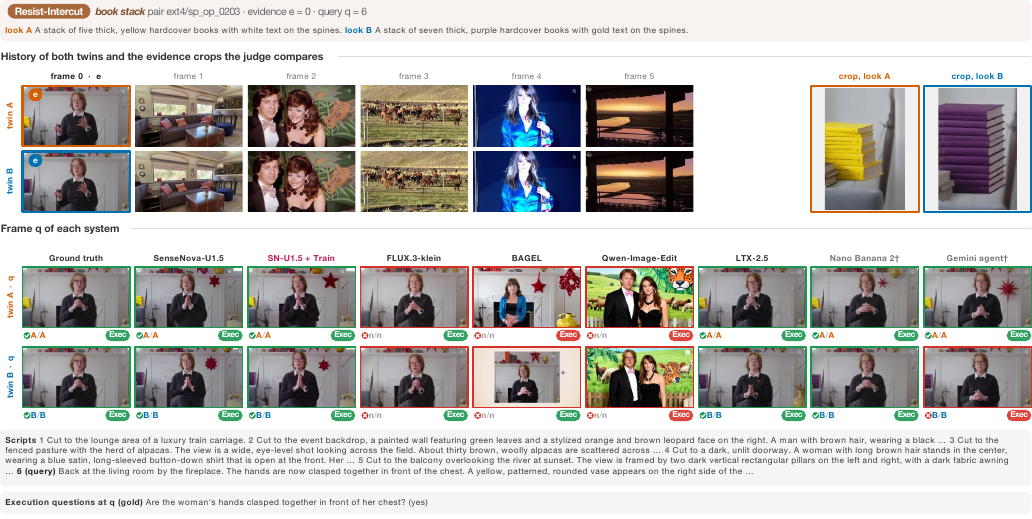}
\caption{\textbf{Resist-Intercut example.} Layout and selection as in \cref{fig:viz-sub0}.}
\label{fig:viz-sub4}
\vspace{12pt}
\includegraphics[width=1.000\linewidth]{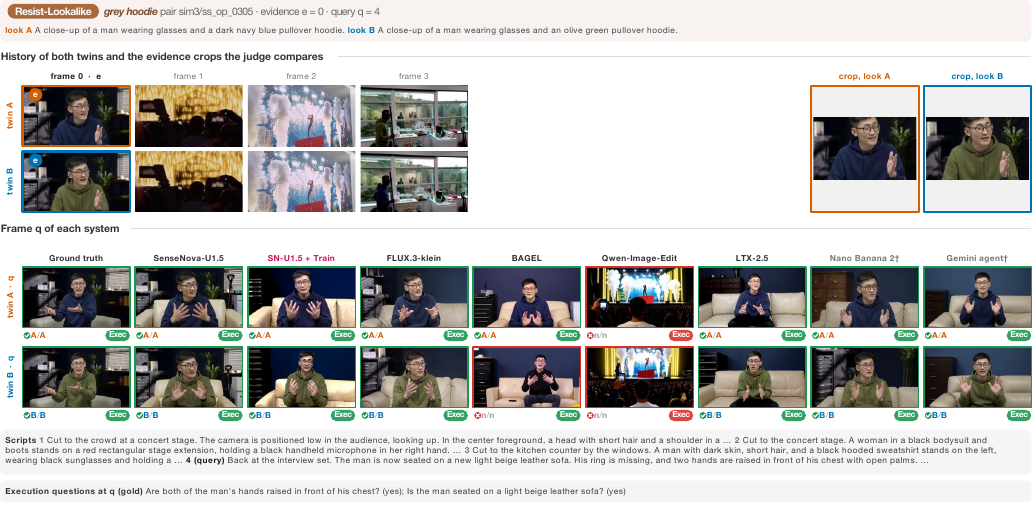}
\caption{\textbf{Resist-Lookalike example.} Layout and selection as in \cref{fig:viz-sub0}.}
\label{fig:viz-sub5}
\end{figure*}

\begin{figure*}[p]
\centering
\includegraphics[width=0.897\linewidth]{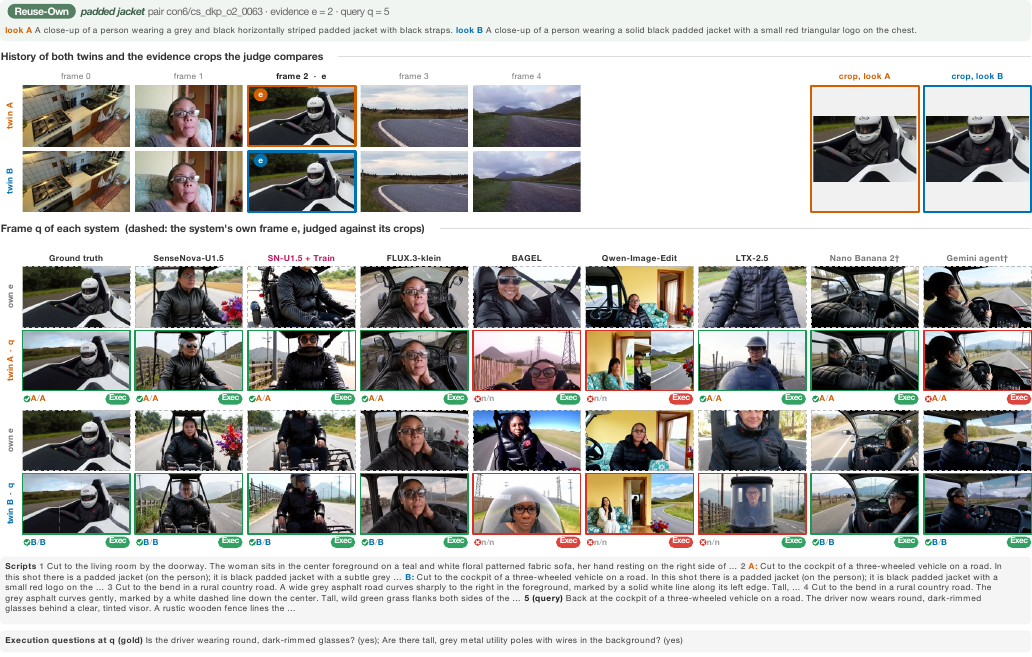}
\caption{\textbf{Reuse-Own example.} For Reuse the twins differ in the script of $e$ and in the frame the model draws from it, so each column shows the model's own $e$ dashed above its output and is judged against crops from it. Layout and selection as in \cref{fig:viz-sub0}.}
\label{fig:viz-sub6}
\vspace{12pt}
\includegraphics[width=0.897\linewidth]{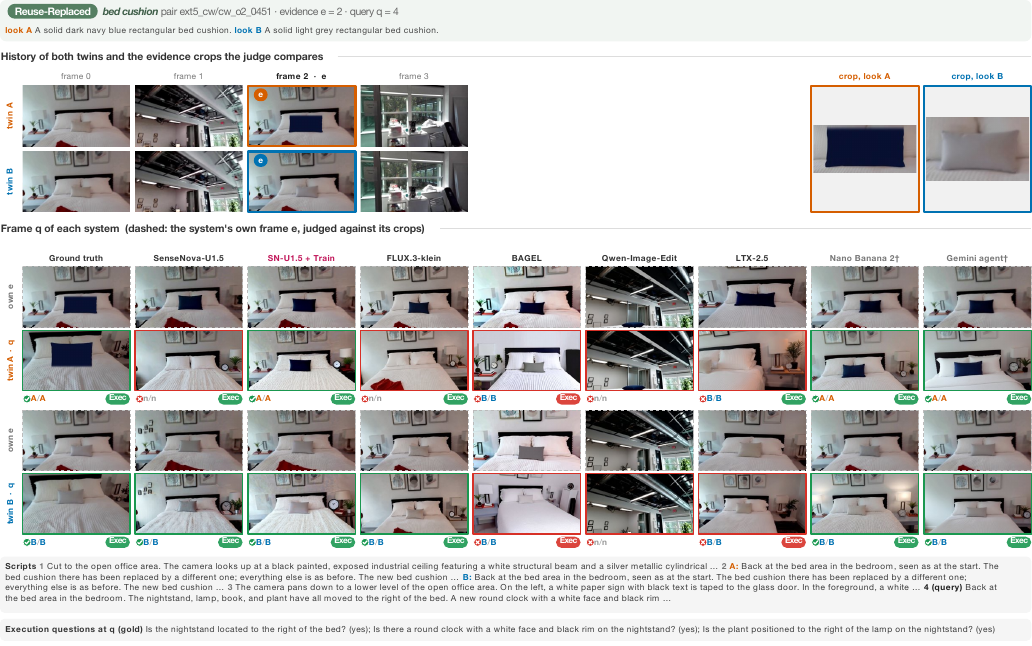}
\caption{\textbf{Reuse-Replaced example.} Layout and selection as in \cref{fig:viz-sub0}.}
\label{fig:viz-sub7}
\end{figure*}

\begin{figure*}[p]
\centering
\includegraphics[width=1.000\linewidth]{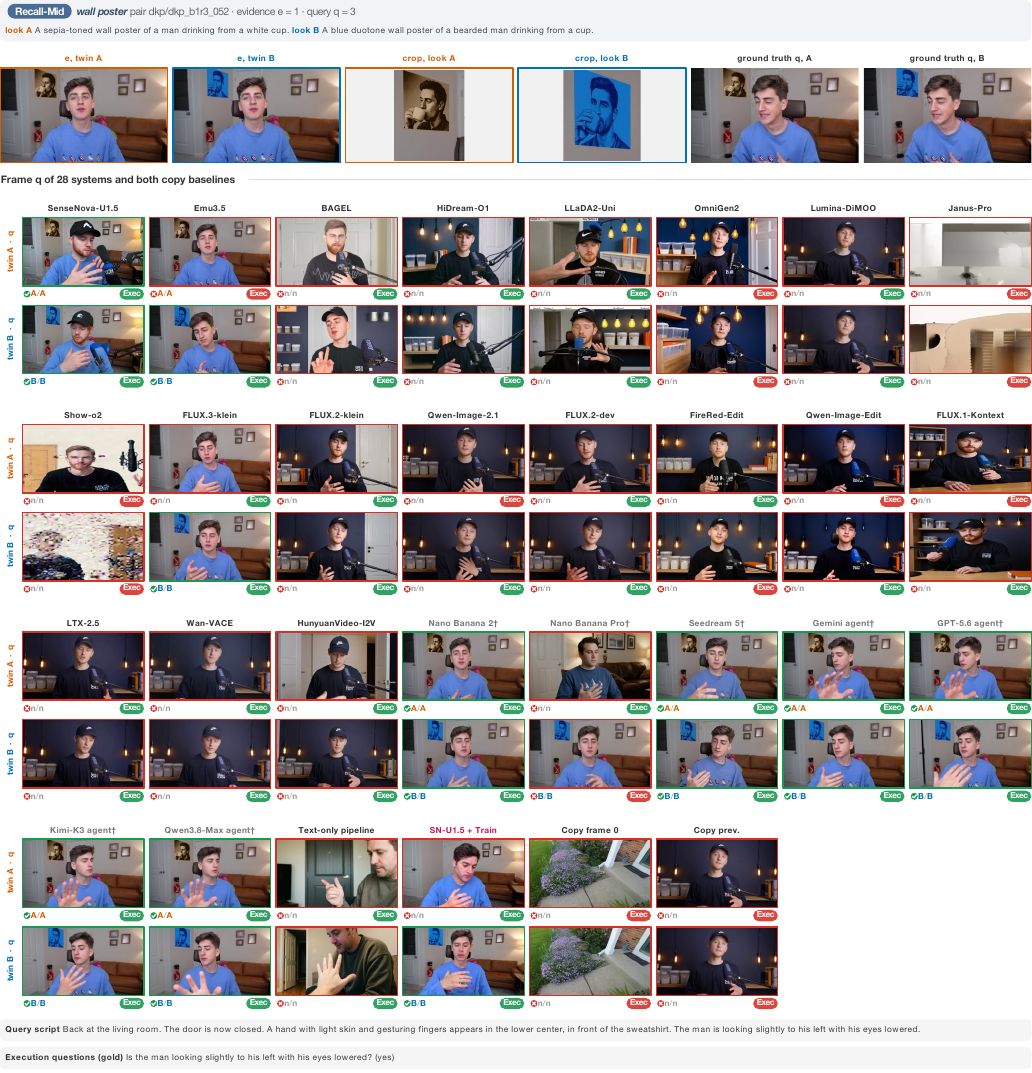}
\caption{\textbf{28 systems on one Recall-Mid pair.} Top: the evidence frame and true query frame of each twin, and the judge's two crops. Below: frame $q$ of each system and of both copy baselines for both twins, marked as in \cref{fig:viz-sub0}. $^\dagger$ marks a closed planner or generator.}
\label{fig:viz-grid0}
\end{figure*}

\begin{figure*}[p]
\centering
\includegraphics[width=1.000\linewidth]{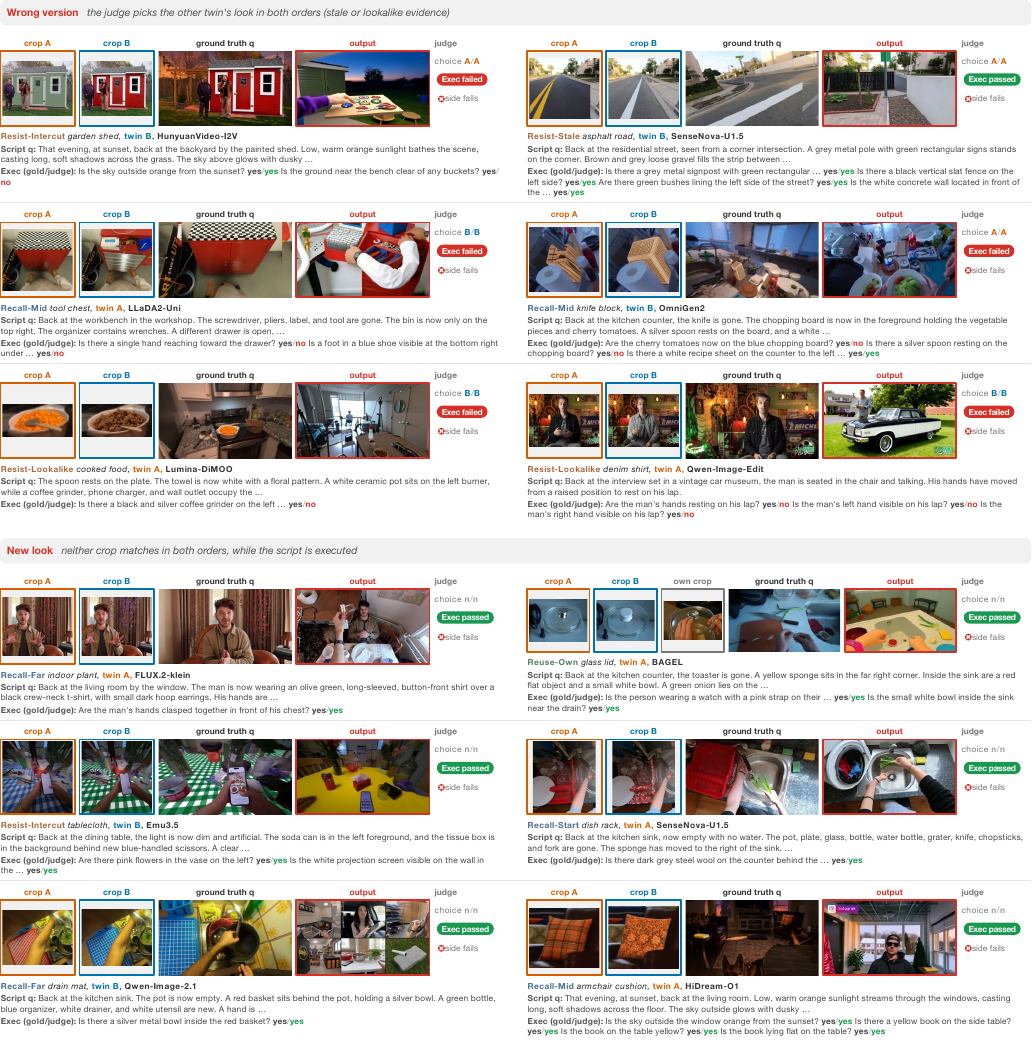}
\caption{\textbf{Failure modes} on phase~A, one twin side per case. Each case shows the evidence crops, for Reuse also the crop of the system's own $e$, the true query frame and the output. Beside them are the judge's answers in both crop orders, the Exec questions with gold and judge answers, wrong ones in red, and the side result. Cases were chosen automatically.}
\label{fig:viz-fail0}
\end{figure*}

\begin{figure*}[p]
\centering
\includegraphics[width=1.000\linewidth]{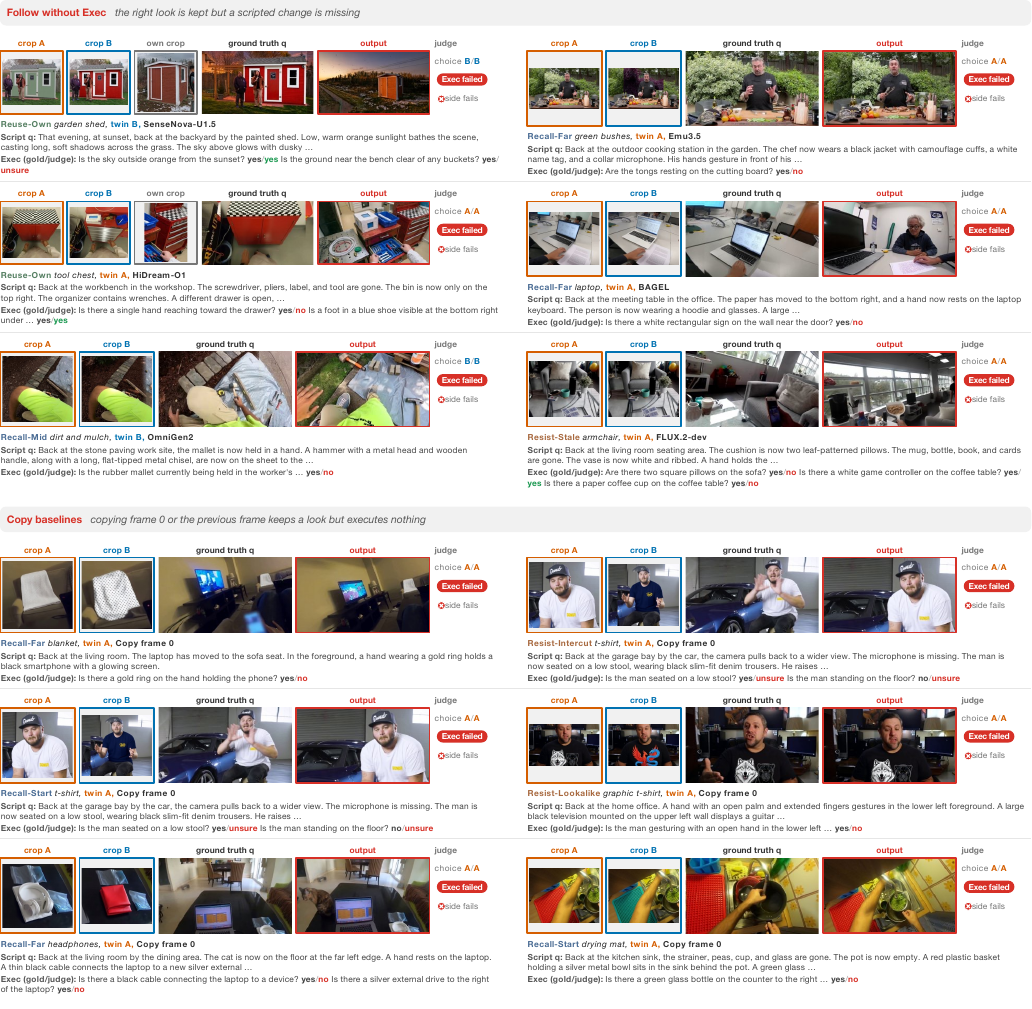}
\caption{\textbf{Failure modes, continued.} Layout and selection as in \cref{fig:viz-fail0}.}
\label{fig:viz-fail1}
\end{figure*}

\begin{figure*}[p]
\centering
\includegraphics[width=1.000\linewidth]{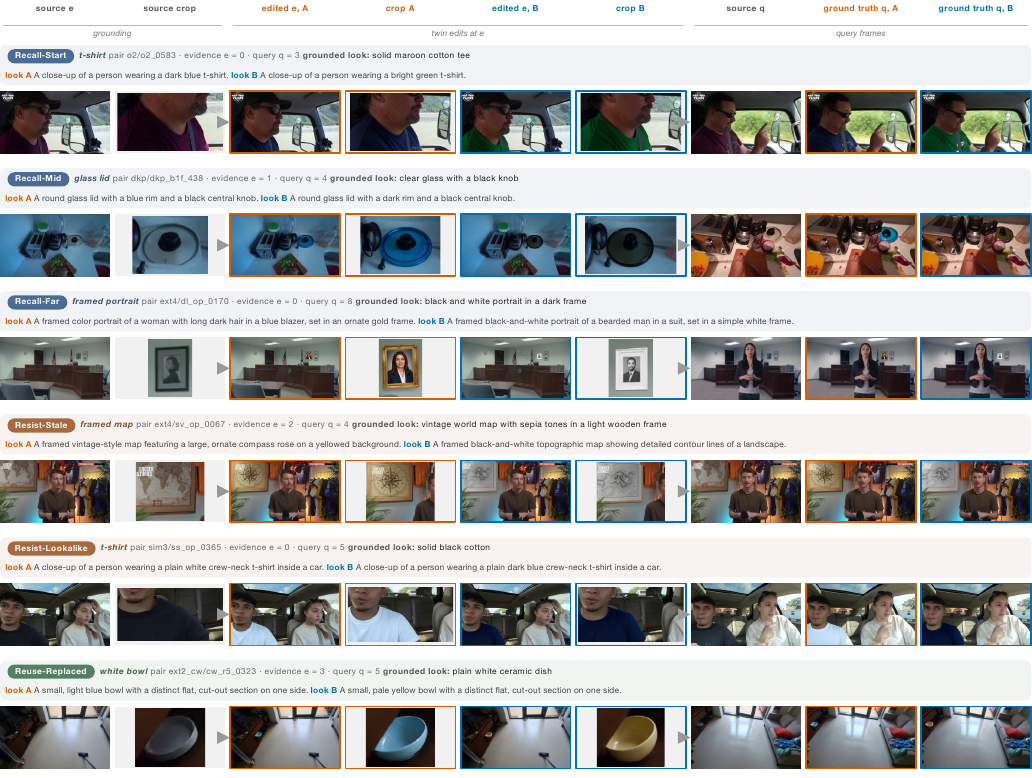}
\caption{\textbf{Twin construction} \crefp{sec:bench}, one pair per subtype. Each row shows the source $e$ with its grounded crop, the two edited copies of $e$ with the judge's crops, the source $q$ and the two true query frames. The text lists the grounded look and looks A and B. In Reuse-Replaced the twins share the opening frame and the replacement is drawn at $e$.}
\label{fig:viz-build}
\end{figure*}

\begin{figure*}[p]
\centering
\includegraphics[width=1.000\linewidth]{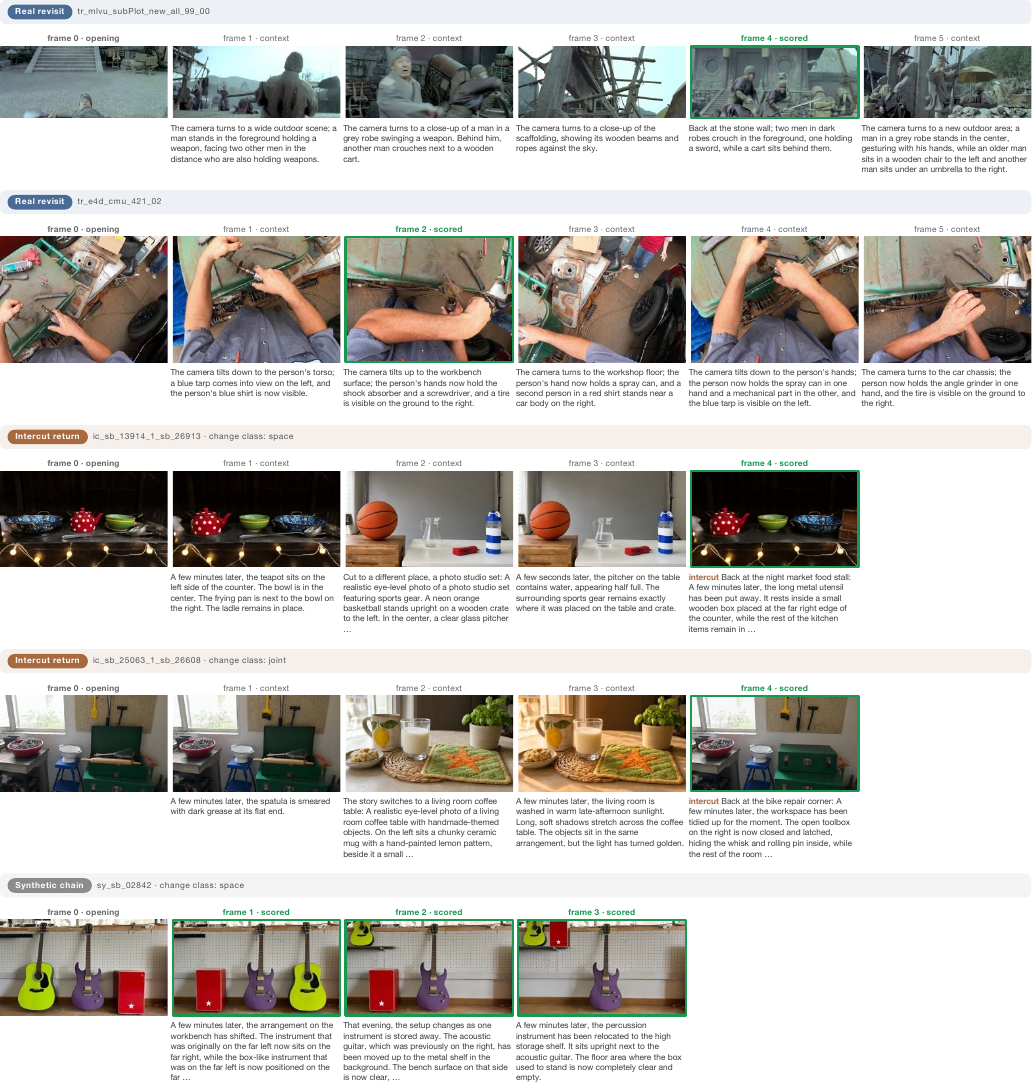}
\caption{\textbf{\train examples}, two random chains per track. Chains have four to six frames, and verified edits have two. Frame 0 opens the chain and each later frame is the target of one script. Frames marked scored, with a green border, enter the loss.}
\label{fig:viz-train0}
\end{figure*}

\begin{figure*}[p]
\centering
\includegraphics[width=0.993\linewidth]{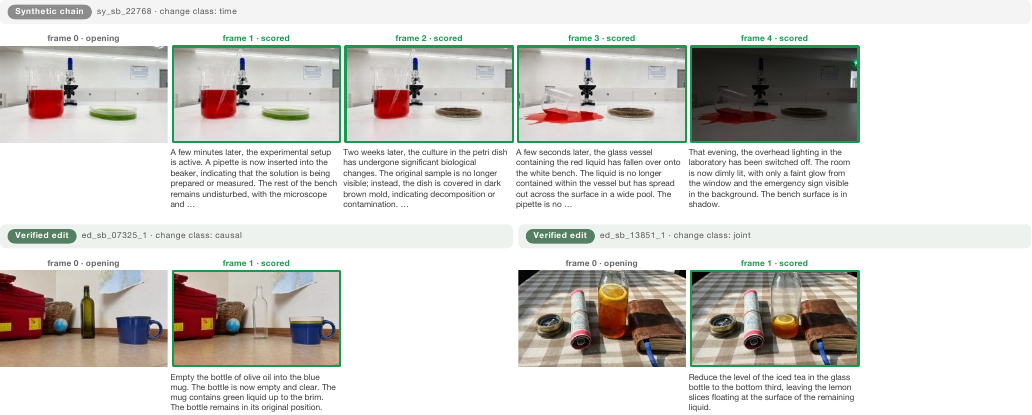}
\caption{\textbf{\train examples, continued.} Layout and selection as in \cref{fig:viz-train0}.}
\label{fig:viz-train1}
\vspace{12pt}
\includegraphics[width=0.993\linewidth]{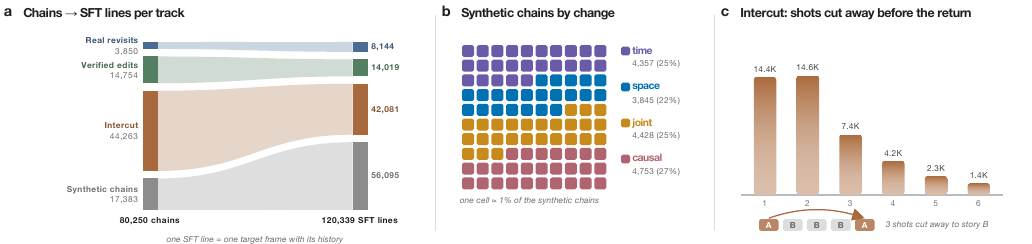}
\caption{\textbf{\train.} \textbf{a} Chains and supervised targets per track. \textbf{b} Synthetic chains per
change class. \textbf{c} Number of shots between leaving and returning to a storyline in intercut chains.}
\label{fig:train}
\end{figure*}

Every pair on the following pages was selected by a fixed rule on the judge's answers, the edit leak
\crefp{app:build} and the source set, without inspecting any image. Each subtype page shows the real-video pair with
the lowest edit leak among those that the ground truth and SenseNova-U1.5 with \train pass and that 2 to 6 of the
eight systems shown pass. Unless noted, pairs come from phase~A, with twin A in \textcolor{lookA}{orange} and twin B
in \textcolor{lookB}{blue}.
\cref{fig:viz-sub0,fig:viz-sub1,fig:viz-sub2,fig:viz-sub3,fig:viz-sub4,fig:viz-sub5,fig:viz-sub6,fig:viz-sub7} give
one pair per subtype with its history, scripts and execution questions, the ground truth and eight systems, and
\cref{fig:viz-grid0} shows 28 systems and both copy baselines on one pair. \cref{fig:viz-fail0,fig:viz-fail1} collect failure modes, \cref{fig:viz-build} shows how the twins of each subtype are
built, and \cref{fig:viz-train0,fig:viz-train1} show chains from the tracks of \train.

\renewcommand{\dblfloatpagefraction}{0.5}
\makeatletter\setlength{\@dblfptop}{0pt plus 1fil}\setlength{\@dblfpbot}{0pt plus 1fil}\makeatother

\end{document}